\documentclass[11pt]{article}

\usepackage{acl}

\usepackage{times}
\usepackage{latexsym}

\usepackage[T1]{fontenc}

\usepackage[utf8]{inputenc}

\usepackage{microtype}

\usepackage{inconsolata}

\usepackage{graphicx}

\title{Tokenization is Sensitive to Language Variation} %

\usepackage{float}
\usepackage{placeins} %
\usepackage{hyperref}
\usepackage{booktabs}
\usepackage[normalem]{ulem}
\usepackage{listings}
\usepackage{fvextra}

\usepackage{times}
\usepackage{latexsym}

\usepackage{graphicx}

\usepackage{pifont}
\usepackage{multirow} %

\usepackage{scalefnt}  %
\makeatletter
\DeclareRobustCommand{\fakesc}[1]{%
  \@fakesc #1\@nil
}
\def\@fakesc#1{%
  \ifx#1\@nil
    \let\next=\relax
  \else
    \@fakescchar{#1}%
    \let\next=\@fakesc
  \fi
  \next
}
\def\@fakescchar#1{%
  \ifnum`#1>64
    \ifnum`#1<91
      \scalebox{0.77}{#1}%
    \else
      \ifnum`#1>96
        \ifnum`#1<123
          \uppercase{\scalebox{0.77}{#1}}%
        \else
          #1%
        \fi
      \else
        #1%
      \fi
    \fi
  \else
    #1%
  \fi
}
\makeatother

\usepackage{tablefootnote} %
\usepackage{amsmath}
\usepackage{subcaption} %
\usepackage{tikz}
\usetikzlibrary{decorations.pathreplacing}
\usepackage{soul}
\usepackage{amssymb}

\author{
    Anna Wegmann\textsuperscript{1}, Dong Nguyen\textsuperscript{1} and David Jurgens\textsuperscript{2} \\
    \textsuperscript{1}Utrecht University, Utrecht, the Netherlands \\
    \textsuperscript{2}University of Michigan, Ann Arbor, the United States \\
    \texttt{\{a.m.wegmann, d.p.nguyen\}@uu.nl}, \texttt{jurgens@umich.edu} \\
  }

\begin{document}
\maketitle
\begin{abstract}
    Variation in language is ubiquitous and often systematically linked to regional, social, and contextual factors. Tokenizers split texts into smaller units and might behave differently for less common linguistic forms. This might affect downstream LLM performance differently on two types of tasks: Tasks where the model should be robust to language variation (e.g., for semantic tasks like NLI, labels do not depend on whether a text uses British or American spelling) and tasks where the model should be sensitive to language variation (e.g., for form-based tasks like authorship verification, labels depend on whether a text uses British or American spelling). We pre-train BERT base models with the popular Byte-Pair Encoding algorithm to investigate how key tokenization design choices impact the performance of downstream models:  the corpus used to train the tokenizer, the pre-tokenizer and the vocabulary size. We find that the best tokenizer varies on the two task types and that the pre-tokenizer has the biggest overall impact on performance. Further, we introduce a new approach to estimate tokenizer impact on downstream LLM performance, showing substantial improvement over metrics like Rényi efficiency. We encourage more work on language variation and its relation to tokenizers and thus LLM performance. 
\end{abstract}

\begin{figure}[t!]
    \centering
    \begin{tikzpicture}[
        every node/.style={font=\small},
        box/.style={draw=orange, thick, dashed, rounded corners, inner sep=10pt, align=center},
        arrow/.style={->, thick},
        diff/.style={text=red}
        ]

        \node[box, minimum width=3.5cm, minimum height=3.5cm] (semantics) at (-1, 0) {};
        \node[above, align=center] at (semantics.north) {\textbf{Tasks Robust to} \\ \textbf{Language Variation}};
        \node[below right, align=center] at (semantics.north west) {\textbf{e.g., NLI}};
        
        \node[align=left] (text1) at (-1.0, 0.5) {\textbf{\textit{guy}} \textit{\textbf{learn}t} \textbf{\textit{guitar}}};
        \node[align=left] (text2) at (-1.0, -0.3) {\textit{A\ \textbf{man}\ }\textit{\textbf{learn}ed}\\ \textit{an\ \textbf{instrument}.}};
        
        \node[box, minimum width=3.5cm, minimum height=3.5cm] (form) at (3, 0) {};
        \node[above, align=center] at (form.north) {\textbf{Tasks Sensitive to} \\ \textbf{Language Variation}};
        \node[below right, align=center] at (form.north west) {\textbf{e.g.,} \\ \textbf{Authorship Verification}};
        
        \node[align=left] (text3) at (3.0, 0.5) {\textit{\uline{g}uy}\ \textit{learn\uline{t}}\ \textit{guitar}\uline{ }};
        \node[align=left] (text4) at (3.0, -0.3) {\textit{\uline{A}\ man\ learn\uline{ed}}\\ \textit{an\ instrument}\uline{.}};

        \draw [decorate, decoration={brace, amplitude=5pt, mirror}, thick] 
    (1.9, -0.9) -- (4.1, -0.9) node[midway, below, yshift=-5pt] {\textbf{Different Author}};
        \draw [decorate, decoration={brace, amplitude=5pt, mirror}, thick] 
    (-2.1, -0.9) -- (0.1, -0.9) node[midway, below, yshift=-5pt] {\textbf{Entailment}};

    \end{tikzpicture}
    \caption{\textbf{Different types of tasks might profit from different tokenizer settings.} We investigate whether the same tokenizer performs well for tasks that require robustness to language variation (i.e., semantic-focused tasks like NLI) and for tasks that require sensitivity to language variation (i.e., form-focused tasks like authorship verification). Intuitively, one needs to look at \textbf{semantic signals} (e.g., ``guitar'' and ``instrument'') for NLI and at \uline{form- or style-based signals} (e.g., ``learn\uline{ed}'' and ``learn\uline{t}'') for authorship verification. %
    }
    \label{fig:semantics_form}
\end{figure}

\begin{table*}[t!]
    \centering
    \small
    \begin{subtable}[h]{0.55\textwidth}
        \centering
        \begin{tabular}{p{1.5cm} p{4.5cm} l}
            \toprule
                & \textbf{Example} & \textbf{Label} \\
            \midrule
            GLUE & How do I hire an ethical hacker? Where can I find ethical hackers? & duplicate \\
            +typo & Wher're cacn I fien ethical hackprs? Hjw fo I hie an ehtical hacker? & duplicate \\
            +dialect & How do I hire me an ethical hacker? Where can I find me ethical hackers?	 & duplicate \\
            \bottomrule
        \end{tabular}
        \caption{\textbf{Tasks requiring robustness to language variation.}}
    \end{subtable}
    \hfill
    \begin{subtable}[h]{0.44\textwidth}
        \centering
        \vspace{\baselineskip}
        \begin{tabular}{p{1.5cm} p{3.8cm}} %
            \toprule
                \textbf{Dataset} & \textbf{Task} \\ %
            \midrule
            AV & Authorship Verification \\%  & 48k \\
            PAN & Author Change Classification \\% & 18k \\
            CORE & Register Classification \\% & 30k \\ %
            NUCLE & Error Classification \\% & 22k \\  %
            Dialect & Dialect Classification \\ %
            \bottomrule
        \end{tabular}
        \caption{\textbf{Tasks requiring sensitivity to language variation.}}
    \end{subtable}
    \caption{\textbf{Evaluation tasks.} We evaluate tokenizers on tasks that require (a) robustness and (b) sensitivity to language variation. For (a), we use GLUE tasks, adding spelling perturbations using \newcite{wang-etal-2021-textflint} and dialect perturbations using \newcite{ziems-etal-2023-multi}. Examples are taken from QQP. For (b), we use a newly compiled selection of tasks  containing our own authorship verification dataset, an author change classification dataset \cite{ayele2024overview}, a register classification dataset \cite{laippala2023register}, a grammatical error correction dataset \cite{dahlmeier-etal-2013-building} and a dialect classification dataset generated with  \newcite{ziems-etal-2023-multi}.\label{fig:tasks}}
\end{table*}

\section{Introduction}

Variation in language is ubiquitous, manifesting in forms such as lexical variation (e.g., vocabulary choice like ``lift'' vs. ``elevator''), spelling variation (e.g., spelling  ``u'' instead of ``you''), and syntactic variation (e.g., word order like ``I gave her the ball.'' and ``I gave the ball to her.''). Such variation is often systematic rather than  random, and is linked to but regional, social, and contextual factors \cite{nguyen-etal-2016-survey,coupland2007style,eckert2012three}.

Tokenizers break up input strings and determine the tokens that are fed into language models. Tokenizers build their vocabulary based on a \textit{fitting corpus} and have a given \textit{vocabulary size}. A \textit{pre-tokenizer} can prioritize or prevent the creation of certain tokens. Depending on these settings, tokenizers might behave differently for linguistic forms that are less common (see \S\ref{sec:tokenizer}). For example, spelling variants of common words might be split into more tokens than their standard spelling \cite{matthews-etal-2024-semantics}, e.g., ``doing'' might be one token, while ``doin'' might be split into ``do'' and ``in''. As a result, LLMs might need to learn to recompose such spellings to represent their meaning, suffering in performance compared to representing words as one token. In contrast, smaller tokens could be useful for recognizing form-based patterns (e.g., distinguish the ``in'' from the ``ing'' ending). %
We ask \textbf{RQ1}: \textit{Do the same tokenizer settings (i.e., fitting corpus, vocabulary size and pre-tokenizer) perform well for two types of downstream tasks: Tasks whose gold labels are \textbf{robust to language variation} (e.g., tasks that involve analyzing the semantic meaning of texts like NLI) and tasks whose gold labels are \textbf{sensitive to language variation} (e.g., tasks centered around the form, style or language variety of texts like authorship verification and dialect classification)?} For example, for semantic tasks like NLI, models should perform equally well, regardless of whether the text is written in Standard American English  or African American English. In contrast, for form-based tasks like authorship verification (i.e., are two texts written by the same author?), models should be able to distinguish British and American spelling. %
Intuitively a model needs to use different signals (see Figure~\ref{fig:semantics_form}) to solve tasks requiring robustness and tasks requiring sensitivity to language variation, and might thus profit from different tokenizer settings. %
We test this by pre-training BERT base models \cite{devlin-etal-2019-bert} with different tokenizer settings of the de-facto standard tokenizer: Byte-Pair Encoding or BPE \cite{sennrich-etal-2016-neural}. %

The evaluation of tokenizers with fast proxy measures --- usually based on how well the tokenizer compresses %
a reference corpus --- offers a popular and cheaper alternative to training larger LMs for comparing different tokenizers. %
However, common proxy measures often do not achieve a high correlation with downstream performance \cite{zouhar-etal-2023-tokenization,cognetta-etal-2024-two,schmidt-etal-2024-tokenization}. One reason could be that such measures are task-agnostic: They predict the same performance for a given downstream corpus, regardless of task or label differences (e.g., in Figure~\ref{fig:semantics_form}, NLI and authorship verification have different labels for the same sentence pair). %
We therefore investigate \textbf{RQ2}: \textit{Can simple task-aware measures based on logistic regression better predict tokenizer downstream performance on variation-robust and variation-sensitive tasks than the common task-agnostic measures Rényi efficiency and Corpus Token Count?}

\textbf{Contributions.} We make the following three contributions:
(1) We provide our code and a selection of tasks (see Table~\ref{fig:tasks}) to evaluate models on tasks requiring sensitivity and robustness to English language variation.\footnote{%
\url{https://github.com/nlpsoc/Tokenization-Language-Variation}
}
(2) We find that the impact of tokenizers on downstream performance varies depending on whether a task requires robustness or sensitivity to language variation, and that the pre-tokenizer has the biggest influence on downstream performance across task types. However, aggregated performance differences remain small and highlight the need for future work to investigate different types of language variation individually.
(3) Further, we find that logistic regression performance has a higher correlation with BERT downstream performance than metrics like {Corpus Token Count} or Rényi efficiency \cite{zouhar-etal-2023-tokenization}. 
We provide practical suggestions to evaluate and build better tokenizers. 
With this work, we hope to encourage more work on language variation and its relation to tokenizers to build better, fairer and more robust LLMs from the ground up.

\section{Related Work}

    Next to Byte-Pair Encoding \cite{sennrich-etal-2016-neural}, there are several other subword (and other) tokenizers \cite{mielke2021between}. %
    For example, character and byte-based tokenizers have been argued to be more robust to spelling variations \cite{mielke2021between, libovicky-etal-2022-dont, xue-etal-2022-byt5}. In this study we focus on BPE as it has been the most common tokenization algorithm in recent LLMs (e.g., Llama 3, Mixtral, DeepSeek v3 and GPT-4). 

    There is no universally agreed-upon standard to evaluate tokenizers. Tokenizers have been evaluated \textit{intrinsically} (i.e., without training larger LMs) and \textit{extrinsically} (i.e., considering the performance of larger LMs pre-trained with the considered tokenizer). Common intrinsic methods include: the average number of subwords produced per word and correlated measures like Corpus Token Count \cite{rust-etal-2021-good, Scao2022BLOOMA1, ali-etal-2024-tokenizer,galle-2019-investigating,schmidt-etal-2024-tokenization}, %
     information theoretic measures like Shannon entropy or Rényi efficiency \cite{zouhar-etal-2023-tokenization} and
     morphological alignment \cite{gow-smith-etal-2022-improving, uzan-etal-2024-greed}. 
    Extrinsic measures include LM perplexity \cite{shliazhko-etal-2024-mgpt,zevallos-bel-2023-hints,gowda-may-2020-finding}, cross-entropy loss \cite{rajaraman2024toward}, computational training cost \cite{ali-etal-2024-tokenizer} and downstream task performances \cite{schmidt-etal-2024-tokenization,ali-etal-2024-tokenizer}. 
    It is computationally infeasible to train SOTA LLMs end-to-end for each version of the tokenizer one wants to evaluate. Recent work tackled this by training ``smaller'' generative language models  with 350M--2.5B parameters for each tokenizer \cite{schmidt-etal-2024-tokenization,ali-etal-2024-tokenizer}. However, even such smaller models still can take several days to train. 
    We measure downstream task performance on models with 110M parameters taking less than 15 GPU hours to train per model. %

    Tokenizer algorithms and settings can affect an LLM's performance, for example,
    on tasks including numbers like arithmetic \cite{thawani-etal-2021-representing, wallace-etal-2019-nlp},
    on tasks including domain specific vocabulary or jargon like coding or medicine \cite{gu2021domain, dehaerne2022code, zan-etal-2023-large},
    on different scripts and languages \cite{petrov_unfairnesstokenizers,rust-etal-2021-good,limisiewicz-etal-2023-tokenization,ahia-etal-2023-languages,  velayuthan-sarveswaran-2025-egalitarian}
    and when translating between languages \cite{galle-2019-investigating,libovicky-etal-2022-dont,zhang-etal-2022-robust}.
    To the best of our knowledge monolingual tokenizers have been underinvestigated in relation to language variation. Monolingual tokenizers and tokenizer settings have recently been investigated on broader selections of NLU tasks \cite{schmidt-etal-2024-tokenization,ali-etal-2024-tokenizer}, presumably with the underlying assumption that for a given language like English, there exists a best tokenizer for most, if not all, tasks. We investigate this assumption for two systematically different types of tasks: tasks requiring robustness and tasks requiring sensitivity to language variation.

\section{Tokenizer Settings} \label{sec:tokenizer}

Tokenizers break up texts into smaller units. These units are fed into the language model as input. We investigate different variations of the most popular tokenization algorithm: Byte-Pair Encoding or BPE \cite{sennrich-etal-2016-neural}. BPE is a subword tokenization algorithm (i.e., it breaks rare words down into subword units, bottoming out in bytes in the worst case). The vocabulary is built iteratively: It starts out with a base vocabulary of distinct %
bytes %
and merges them based on token frequency in the fitting corpus until the desired vocabulary size is reached.
We vary the algorithm on three parameters: (1) the vocabulary size, (2) the pre-tokenizer and (3) the fitting corpus. %

\subsection{Vocabulary Size} \label{sec:vocab-size}

Common vocabulary sizes range from 30k to 64k in monolingual models, to about 128k (Llama~3) and 200k (GPT-4o) in recent multilingual models. Previous work tested vocabulary sizes between 32k and 100k on NLU tasks, finding only small differences in performance between vocabulary sizes for the same model size \cite{ali-etal-2024-tokenizer,schmidt-etal-2024-tokenization}. However, vocabulary size might still play a role when dealing with language variation. When using lower range vocabulary sizes, less common words or spelling variants might not be represented with one token, influencing a word's position in the embedding space \cite{matthews-etal-2024-semantics}. %
Character-based models like ByT5 %
show better robustness to spelling variation than subword models with larger vocabulary sizes \cite{libovicky-etal-2022-dont,xue-etal-2022-byt5,tay2021charformer}. Similarly, tokenizers using low-range (in the extreme case character-level) vocabulary sizes might be more robust to spelling variation, as, for example, swapping of characters will not drastically alter the character-based segmentation of a word. %
We experiment with the following vocabulary sizes: 500 and 4k (low-range vocabulary size), 32k and 64k (mid-range vocabulary size) and 128k (high-range vocabulary size). %

\subsection{Pre-Tokenizer}

    Pre-tokenizers split the fitting corpus into word-like units \cite{mielke2021between}, so-called ``pre-tokens'' (e.g., ``I'm fine.'' $\rightarrow$ ``I'', ``'m'', ``fine'', ``.'') which are then further tokenized independently. That is, in building its vocabulary, the BPE algorithm cannot merge tokens that cross pre-token boundaries (e.g., ``I'm''). Pre-tokenizers are usually expressed with regular expressions that include hard-coded knowledge about a language or script \cite{velayuthan-sarveswaran-2025-egalitarian}, e.g., whitespaces separate words in Latin script and ``'m'' is a contraction in English. %
    The most common English pre-tokenizers are more or less elaborate variations of splitting texts based on white\-spaces. For example, {GPT-2}'s pre-tokenizer additionally separates different \textit{Unicode character categories} (e.g., letters, numbers and punctuation) but leaves single leading whitespaces attached to words \cite{radford2019gpt2}. 
    
    Pre-tokenizers influence compression efficiency \cite{radford2019gpt2}, NLU downstream performance \cite{schmidt-etal-2024-tokenization} and arithmetic performance \cite{lee2024digitstodecisions}. %
    However, pre-tokenizers have not been investigated for tasks that require sensitivity to language variation. 
    Pre-tokenizers might be important for handling words that merge {Unicode character categories}, for example, when letters are substituted with numbers (e.g., ``2day'' or ``c000l'', see \newcite{eger-etal-2019-text} for more). If a pre-tokenizer generally splits off numbers and letters, such words can never become part of the vocabulary. 
    Further, not splitting on whitespace might better represent syntactic variation by including frequent phrases like ``of the''. Frequent compositional phrases are processed faster by humans \cite{arnon2010more}, and building tokenizers that align closer to such cognitive processes (e.g., \citealp{yang-etal-2020-less}) might also be beneficial for semantic tasks. %

    We compare five different pre-tokenizers: (a)~not using a pre-tokenizer (\fakesc{no}), (b)~isolating whitespaces (\fakesc{ws}), (c)~leaving single whitespaces attached to words (\fakesc{\_ws}),
    (d) Llama 3's pre-tokenizer (\fakesc{llama3}) %
    and (e)~GPT-2's widely used pre-tokenizer (\fakesc{gpt2}). %
    \fakesc{llama3} and \fakesc{gpt2} can be understood as extensions of the \fakesc{\_ws} pre-tokenizer. Among others, the \fakesc{llama3} pre-tokenizer splits off Unicode character categories (e.g., punctuation) but leaves one leading non-letter character attached to letters (e.g., 'm). The \textsc{gpt2} pre-tokenizer is similar to \textsc{llama3}, but usually separates Unicode character categories more. %
    See Appendix~\ref{app:tokenizer-pretokenizer} for the regular expressions describing the  \fakesc{ws}, \fakesc{\_ws}, \fakesc{llama3} and \fakesc{gpt2} pre-tokenizer. %

\subsection{Fitting Corpus} \label{sec:fitting-corpus}

Sub-word tokenizers construct their vocabulary based on a fitting corpus, adding tokens based on the distribution of tokens in that corpus. %
Lexical, morphological and spelling variation are all intuitively connected to the fitting corpus. If a fitting corpus does not show a variation, it can not be part of a single token in the vocabulary.

We sampled fitting corpora with a size of approximately 1.5 billion words. \textbf{PubMed} was randomly sampled from The Pile's PubMed Abstracts \cite{gao2020pile}. The PubMed %
Abstracts consist of 30M abstracts from biomedical articles.
\textbf{Wikipedia} was randomly sampled from Wikipedia articles from a snapshot from June 1st, 2023 after the plain text of articles was extracted.\footnote{using \url{https://github.com/LorcanJConnolly/WikiTextExtractor}.} %
\textbf{Twitter} was sampled from the Decahose Twitter stream\footnote{Decahose provided access to 10\% of real time tweets sampled by Twitter.} throughout the year 2021, with queries on almost every day of the year 2021. %
We only select English tweets based on Twitter's internal language identification system. %
We exclude retweets. %
\textbf{Miscellaneous} was sampled from a variety of domains with no overlap with the other fitting corpora. It includes Reddit, literature sources (fanfictions and books), news articles and comments, question answering websites, reviews, mails, transcripts, blogs, Common Crawl, scientific articles, code and mathematical formulas. See details in Appendix \ref{app:mixed-dist}.

We expect the fitting corpora to differ in the lexical, syntactic and lexical variation they exhibit. For example, PubMed probably contains less spelling and lexical variation than Twitter. Intuitively, a  tokenizer constructed on PubMed should thus be less capable in representing stylistic variation than one constructed on Twitter. However, it remains unclear how important the fitting corpus composition is. 
\newcite{zhang-etal-2022-robust} investigate different compositions of the fitting corpus in the multilingual setting. They find a surprising robustness to language imbalance in the fitting corpus for languages sharing the same script.

\section{Evaluation Tasks} \label{sec:tasks}

We compare tokenizers on classification tasks that require robustness to language variation (\S\ref{NLU}) and tasks that require sensitivity to language variation (\S\ref{language-variation}). Models solving these two types of tasks should need to make use of more semantic and form-based signals respectively and might have different requirements for a tokenizer (cf. Figure~\ref{fig:semantics_form}). We select tasks that strike a balance between being sufficiently challenging and staying within the capabilities of our pre-trained \textsc{BERT} models. See an overview of selected tasks in Table \ref{fig:tasks}.

\subsection{Tasks Robust to Language Variation} \label{NLU}

First, we evaluate tokenizers on semantic tasks where the gold label is robust to language variation. We use GLUE \citep{wang-etal-2018-glue}, a standard NLP benchmark that was also used to evaluate BERT at its introduction \cite{devlin-etal-2019-bert}, and is within the capabilities of our pre-trained BERT models. %
We compare tokenizers on the following four GLUE tasks: %
SST-2 (sentiment classification), QQP (paraphrase classification), MNLI and QNLI (NLI tasks). 
For details on our task selection see Appendix~\ref{App:NLU}. 
Ideally, a robust tokenizer performs consistently across all versions of \textbf{GLUE}:  the original, primarily written in Standard American English, the spelling-transformed \textbf{GLUE+typo} and the dialect-transformed \textbf{GLUE+dialect}.

\textbf{GLUE+typo.} We use \texttt{textflint} \cite{wang-etal-2021-textflint} to introduce simulated typos and spelling errors to our tasks, similar to \newcite{libovicky-etal-2022-dont}. It uses random character swapping and a list of common spelling errors.

\textbf{GLUE+dialect.} We use \texttt{Multi-VALUE} \cite{ziems-etal-2023-multi} to introduce simulated dialectal variation to our GLUE tasks. \texttt{Multi-VALUE} makes use of 189 dialectal perturbation rules \citep{ziems-etal-2023-multi} operationalized based off of eWAVE \cite{ewave}. %
For each example in the GLUE tasks, we randomly choose between transformations to the following dialects: Appalachian English, Chicano English, Colloquial Singapore English, Indian English and Urban African American English. %

\subsection{Tasks Sensitive to Language Variation} \label{language-variation}

Second, we evaluate tokenizers on tasks that are sensitive to language variation. The gold label for such tasks should be sensitive to stylistic or form-based signals. %

\textbf{AV.} Models performing \textbf{a}uthorship \textbf{v}erification (i.e., are two texts written by the same author?) usually need to be sensitive to the different styles and forms used by different authors \citep{zhu-jurgens-2021-idiosyncratic,wegmann-etal-2022-author,wang-etal-2023-authorship}. Past work has found authorship verification to be sensitive to tokenization, with significant gaps between the BERT and RoBERTa tokenizers \citep{zhu-jurgens-2021-idiosyncratic}. Therefore, we curate a new authorship verification dataset of 40.8k train, 2.5k dev and 4.8k test pairs of texts from different domains, similar in distribution to the Miscellaneous corpus (cf.~\S\ref{sec:tokenizer}). Labels are balanced in the test set. See details in Appendix~\ref{app:sadiri}. 

\textbf{PAN.} We use the PAN 2024 Multi-Author Writing Style Analysis task %
to predict whether an author shift occurs between two consecutive paragraphs extracted from Reddit \cite{ayele2024overview}. Specifically, we use the `hard' task, where paragraphs are about the same topic. We sample such that labels are balanced. %
This results in a training set of 18k and a dev set of 4k instances.

\textbf{NUCLE.}
We use the NUCLE 3.3 corpus %
\cite{dahlmeier-etal-2013-building} for multi-label classification of the errors that were made by English learners in a given sentence. NUCLE was annotated by professional English instructors for 27 error types (e.g., verb tense or article use). %
From NUCLE, we take all 22k unique sentences containing errors and sample 5k error-free sentences. We split it into a train (80\%) and dev (10\%) set. %

\textbf{CORE.} 
We use the \textbf{C}orpus of \textbf{O}nline \textbf{R}egisters of \textbf{E}nglish  \cite{laippala2023register} for register classification. Register is one of the most important factors associated with linguistic variation \cite{biber2012register}. We use the 8 main register labels (e.g., spoken or informational description) for multi-class prediction. To increase the occurence of rarer labels, we split long texts and reach a train size of 30k and a dev size of 5k. See Appendix \ref{app:sadiri} for details. %

\textbf{Dialect.} We randomly sample 60k instances from GLUE+dialect (\S\ref{NLU}) and the original GLUE task, to create a dialect classification task with five dialects and Standard American English in the original GLUE texts. We use 50k texts for the train and 5k for the dev set. %

\begin{table*}[t!]
    \centering\small
    \begin{tabular}{ll | ccc |c}
    \toprule
    \textbf{Tokenizer Setting} & \textbf{Model} & \textbf{GLUE} & \textbf{+typo} & \textbf{+dialect}  & \textbf{AVG}\\
    \midrule
     \multirow{3}{*}{\textbf{Fitting Corpus}} 
      & PubMed 
        & 80.8 $\pm$ 0.0
        & \textbf{69.1} $\pm$ 0.2 %
        & 78.6 $\pm$ 0.2 %
        & \textit{76.2} $\pm$ 0.0 \\
      & Wikipedia
        & 80.7 $\pm$ 0.3
        & 68.6 $\pm$ 0.2 %
        & \textbf{79.3} $\pm$ 0.2 
        & \textit{76.2} $\pm$ 0.2 
        \\
      & Twitter 
        & \textbf{81.1} $\pm$ 0.0
        & {69.1} $\pm$ 0.5 %
        & {78.8} $\pm$ 0.1 %
        & {\textbf{76.4} $\pm$ 0.2} 
        \\
    \cmidrule{2-6}
    \multirow{5}{*}{\textbf{Pre-Tokenizer}}
        & \fakesc{no} 
            & 72.1 $\pm$ 1.0
            & 61.6 $\pm$ 0.1 %
            & 70.1 $\pm$ 0.3 %
            & 67.9 $\pm$ 0.3 \\
        & \fakesc{ws}
            & 80.8 $\pm$ 0.4
            & 68.2 $\pm$ 0.3
            & \textbf{79.3} $\pm$ 0.3 
            & \textit{76.1} $\pm$ 0.3 
            \\
        & \fakesc{\_ws} 
            & 80.8 $\pm$ 0.3
            & \textbf{68.9} $\pm$ 0.1  
            & 79.0 $\pm$ 0.2
            & \textbf{{76.2}} $\pm$ 0.1 
            \\
        & \fakesc{llama3} 
            & 80.9 $\pm$ 0.1
            & 68.2 $\pm$ 0.2 
            & {79.0} $\pm$ 0.0 
            & {\textit{76.1}} $\pm$ 0.0 
            \\
        & \fakesc{gpt2} 
            & \textbf{81.3} $\pm$ 0.4 
            & 68.2 $\pm$ 0.2
            & {79.2} $\pm$ 0.4 
            & {{\textbf{76.2} $\pm$ 0.3}} 
            \\
        \cmidrule{2-6}
     \multirow{5}{*}{\textbf{Vocabulary Size}}
        & 500
            & 77.2 $\pm$ 2.5
            & \textbf{70.3} $\pm$ 2.6   
            & 75.6 $\pm$ 2.0 
            & {74.4} $\pm$ 2.4 
            \\
     & 4k 
            & 80.5 $\pm$ 0.8 
            & \textbf{70.3} $\pm$ 0.9 %
            & 78.6 $\pm$ 0.8 %
            & \textbf{76.4} $\pm$ 0.8  
            \\
      & 32k & \textbf{81.3} $\pm$ 0.4 
            & 68.2 $\pm$ 0.2
            & \textbf{79.2} $\pm$ 0.4 %
            & {{76.2} $\pm$ 0.3}
            \\
     & 64k 
            & 80.8 $\pm$ 0.4 
            & 67.6 $\pm$ 0.6 
            & \textbf{79.2} $\pm$ 0.2 
            & {{75.9} $\pm$ 0.4} 
            \\
     & 128k 
            & 78.7 $\pm$ 2.0
            & 64.6 $\pm$ 1.9 %
            & 76.1 $\pm$ 2.6 %
            & 73.1 $\pm$ 2.2
            \\
    \bottomrule
    \end{tabular}
    
    \caption{
        \textbf{Performance on tasks requiring robustness to language variation.} \label{tab:NLU-performance}
        We display BERT performances, averaged on the original four GLUE tasks and their perturbations using spelling mistakes (+typo) and dialectal transformations (+dialect). We provide the mean and standard deviation ($\pm$) over three seeds respectively. We boldface the best performances for each column and investigated setting. For the averaging column (AVG), italics indicate tokenizers with no statistically significant difference from the best-performing tokenizer (cf.~Figure \ref{fig:sig_robust}).
    }
\end{table*}

\section{Modeling}

For each investigated tokenizer (cf.~\S\ref{sec:tokenizer}), we pre-train three BERT models with different seeds. We use encoder instead of decoder models, as encoder models tend to reach higher performance for classification tasks for low parameter settings. This allows us to train models using fewer GPU hours. %

\textbf{Pretraining BERT Models.}   \label{pretrain}\label{sec:pretrain}
We experiment with pre-training tiny BERT models using a token count T close to 3300M that is exponentially bigger than the 4.6M parameters~P, similar to the original BERT papers \cite{devlin-etal-2019-bert,turc2019well}. However, %
we find that for the same compute, using a bigger model size P and fewer tokens T improves the training loss. %
    Chinchilla's scaling law might also hold for smaller encoder models, specifically optimal parameter count could scale with the token size for a fixed compute $P_{OPT}\approx T^{23/27}$ \cite{chinchilla}. %
For the remainder of this work, we use the base BERT model architecture with 110M parameters, initialize all weights randomly and pre-train on 750M tokens sampled in sequences of 512 from the Miscellaneous corpus (\S\ref{sec:fitting-corpus}) and use a batch size of 32 and 45k steps. %
For further details and hyperparameters, see Appendix~\ref{app:pretraining}.  %

\textbf{Fine-Tuning BERT.} Unless otherwise specified, we use 3 epochs, a max sequence length of 128, a batch size of 32 and a learning rate of 2e-5 to perform the classification tasks. We evaluate on the dev set for GLUE. For comparability, we use the same setup for %
tasks requiring sensitivity to language variation. Only for the authorship verification task we use a contrastive training setup, then use the dev set to find an optimal cosine similarity threshold and calculate accuracy on the test set.

\section{Results} \label{sec:result}

We show the performance of fine-tuned BERT models on tasks that require robustness to language variation in Table \ref{tab:NLU-performance} and tasks that require sensitivity to language variation in Table \ref{tab:Style-performance}. 
When we investigate a specific setting (e.g., the fitting corpus in the first three rows in Table \ref{tab:NLU-performance}), we only change that setting and leave the other at their ``default settings'' to ensure comparability and isolate the effect of each individual setting without exhaustive testing of all possible combinations. We use the following default values for the three settings: the miscellaneous fitting corpus, the \fakesc{gpt2} pre-tokenizer and a vocabulary size of 32k. 

\begin{figure}[t!]
    \centering
    \includegraphics[width=0.32\linewidth]{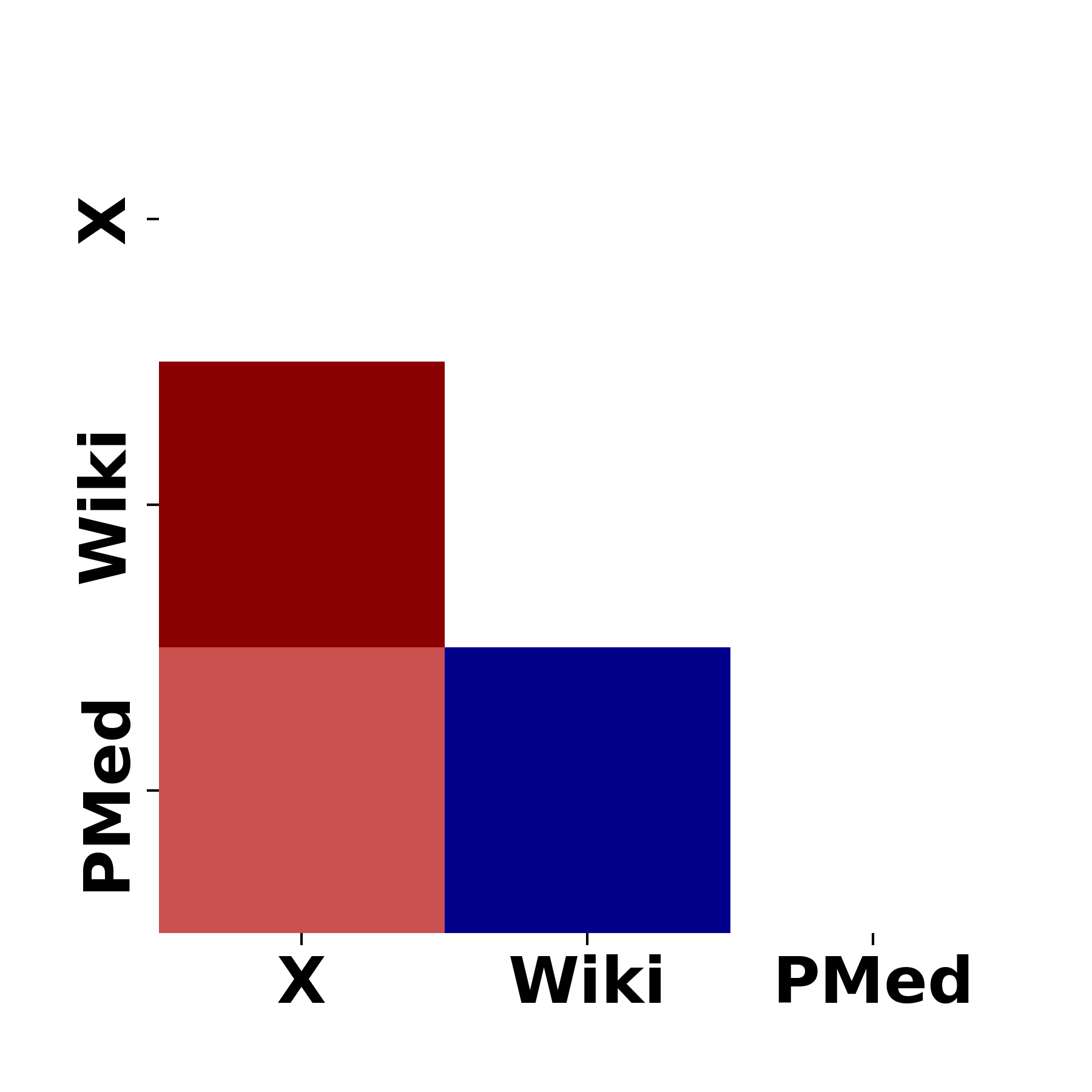}
    \includegraphics[width=0.32\linewidth]{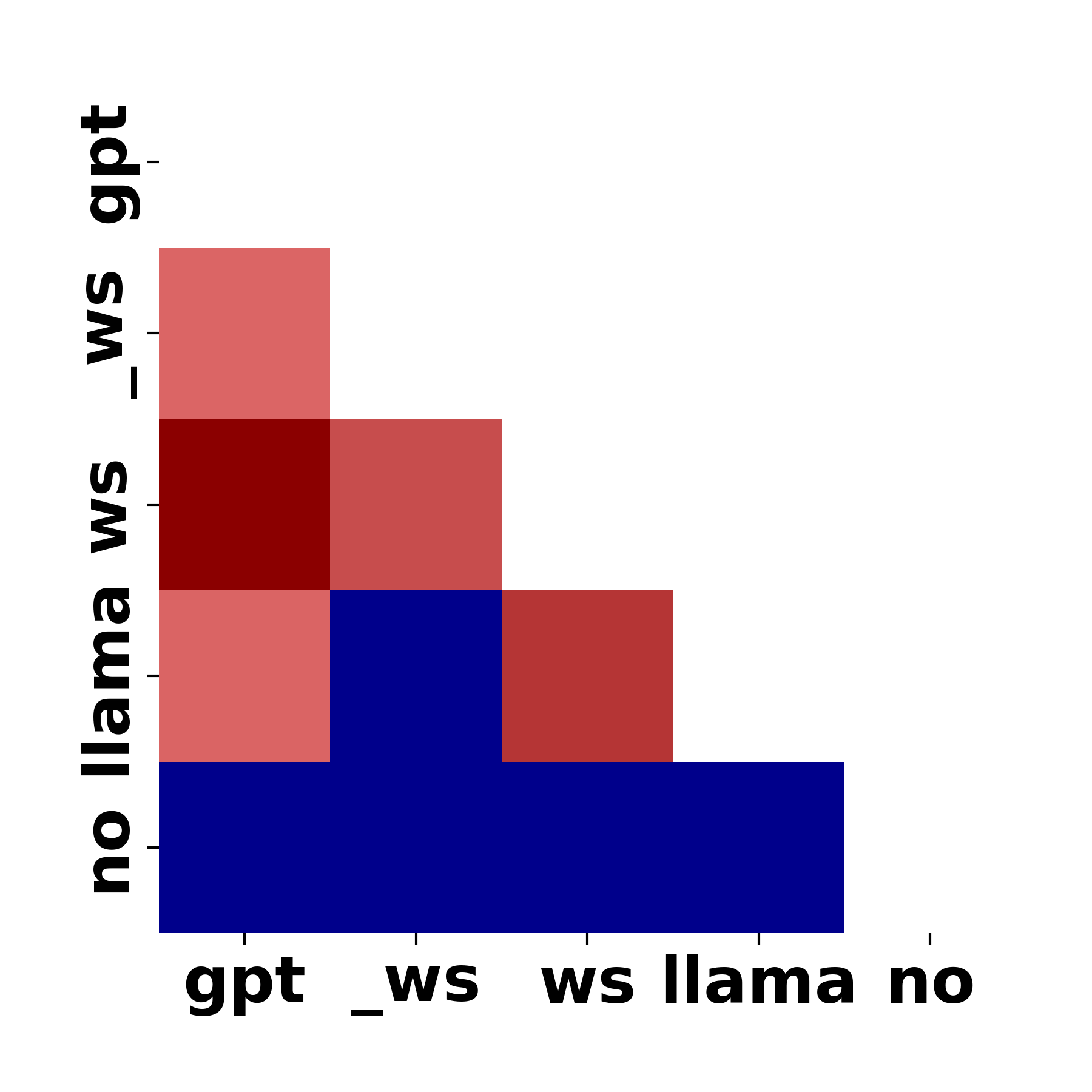}
    \includegraphics[width=0.32\linewidth]{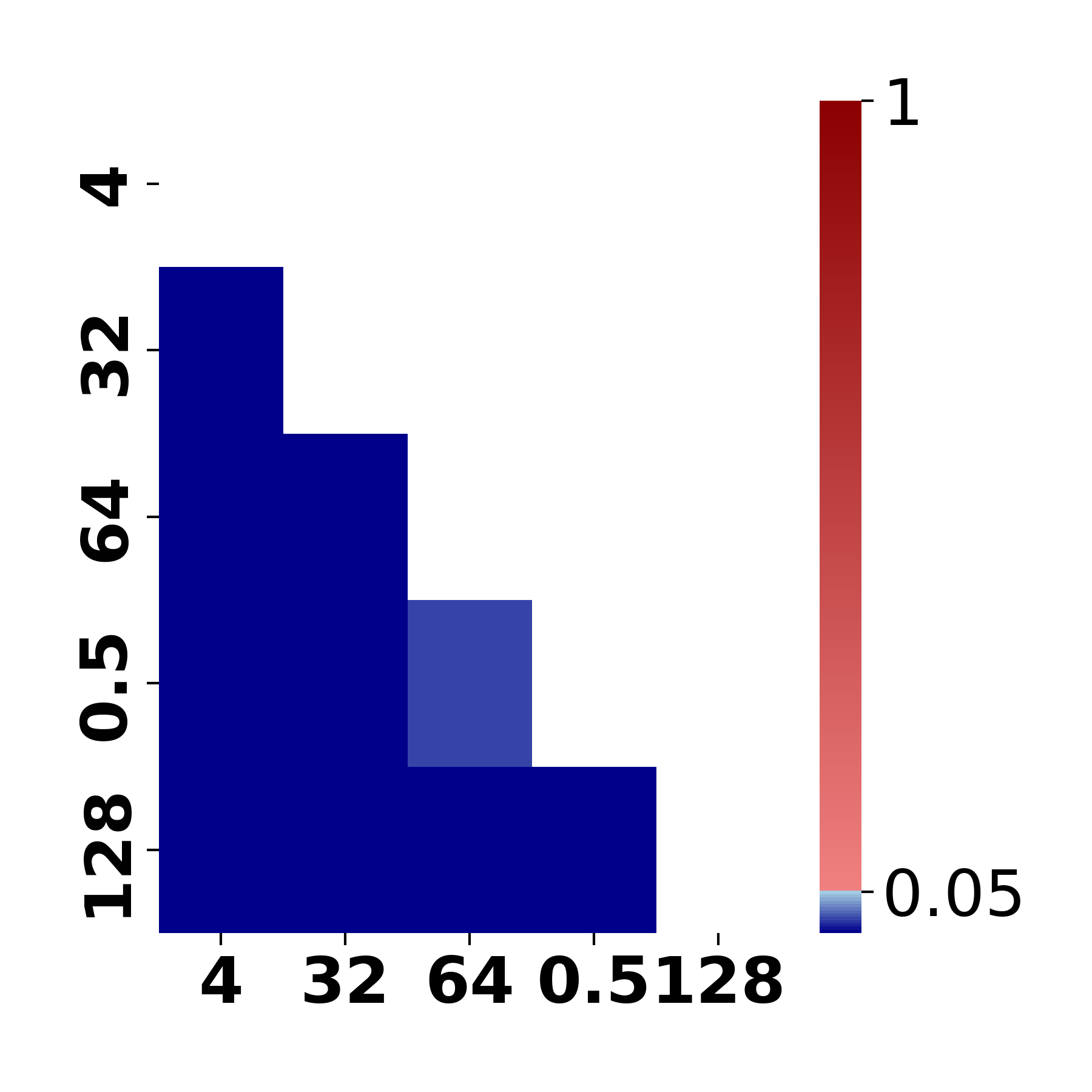}
    \includegraphics[width=0.32\linewidth]{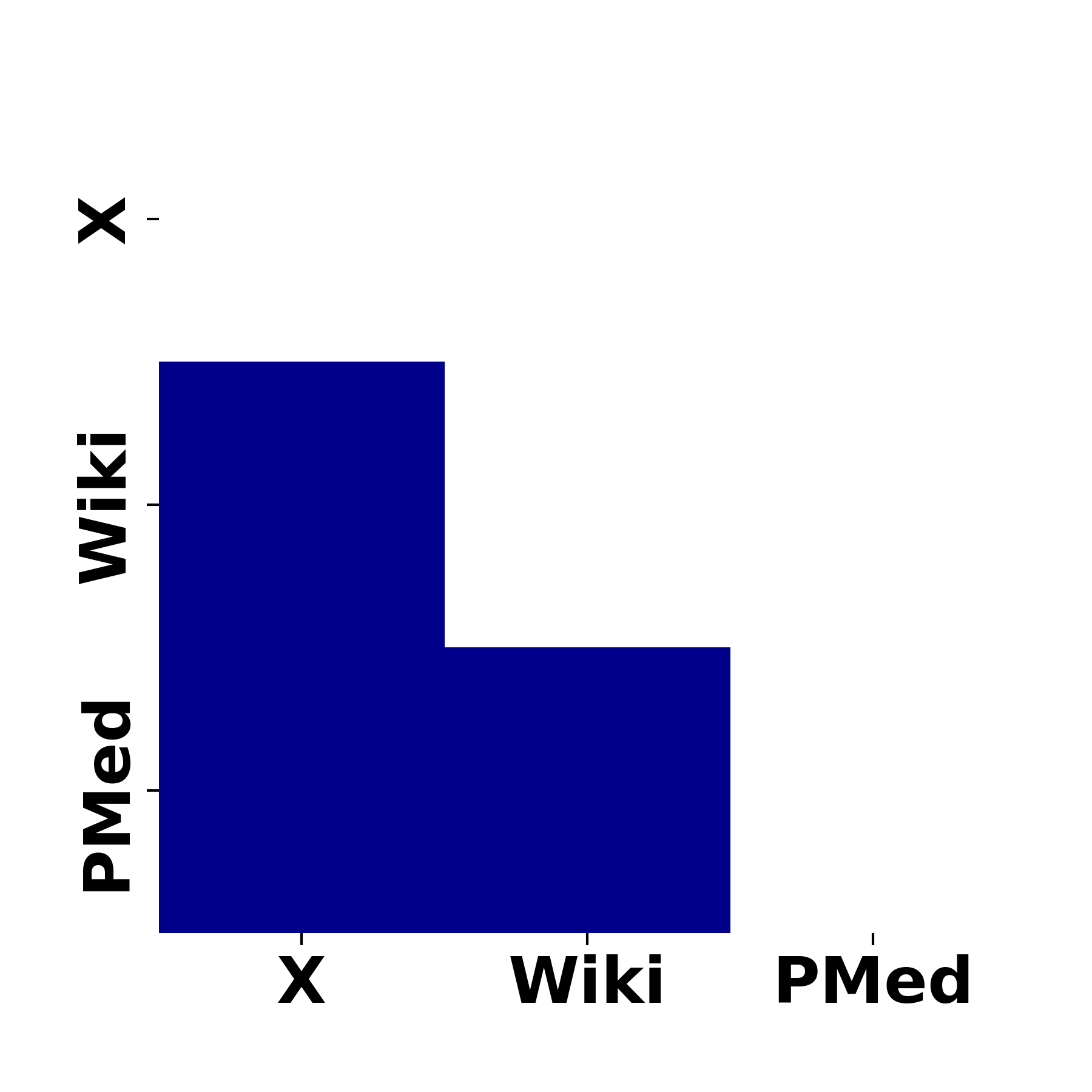}
    \includegraphics[width=0.32\linewidth]{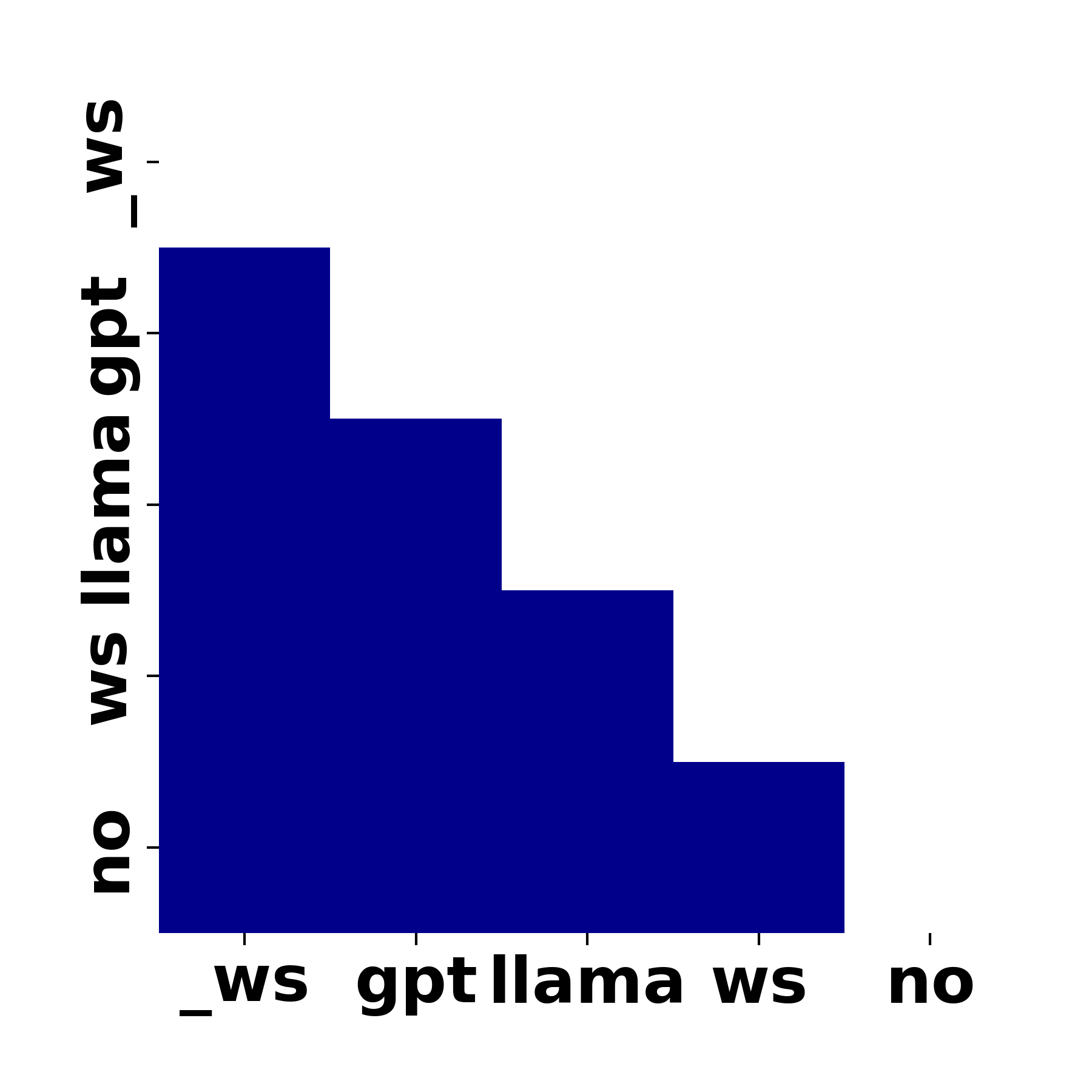}
    \includegraphics[width=0.32\linewidth]{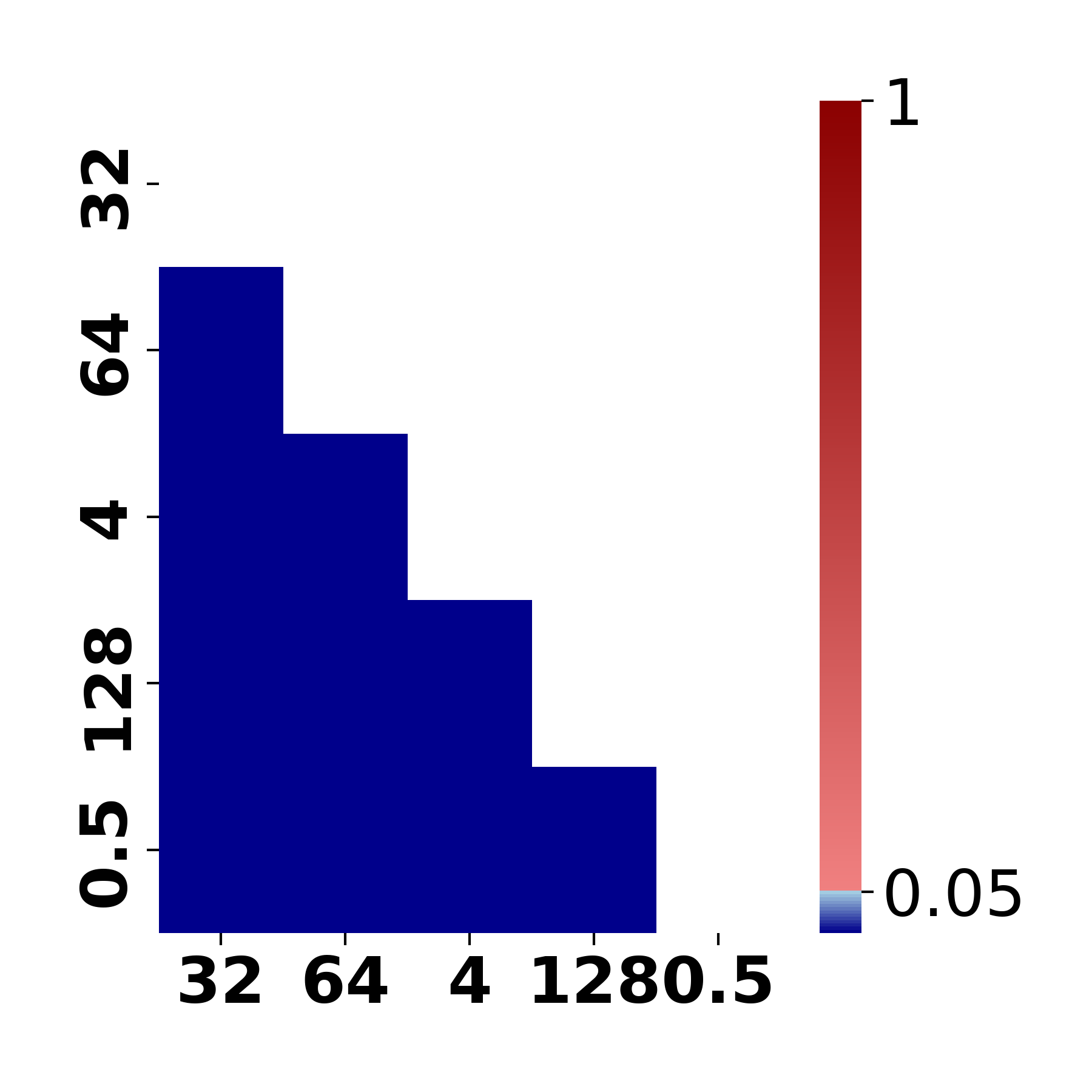}
    \caption{\textbf{Pairwise significance testing of models.}
    We use pairwise \citet{mcnemar1947note}'s test to test whether there is a significant difference between classifications done by models trained with different tokenizers on the tasks requiring robustness to language variation (first row) and tasks requiring sensitivity to language variation (second row). Tokenizers are sorted by mean performance. Blue colors show statistical significance, while red colors are above the 0.05 threshold.
    }
    \label{fig:sig_robust}
\end{figure}

Note that the performance differences tend to be relatively small, which is consistent with previous work comparing different tokenizer algorithms on downstream tasks \cite{ali-etal-2024-tokenizer, schmidt-etal-2024-tokenization}. 
To ensure significance, we compute the pairwise \citealp{mcnemar1947note}'s test %
for the pre-tokenizer, fitting corpus and vocabulary size settings, see Figure \ref{fig:sig_robust}. For the significance testing, we consider classifications by models with the same settings but different seeds to be stemming from the same rater. We use the Bonferroni correction \cite{bonferroni} for our total of 26 tests.

\textbf{RQ1: Tokenizer settings perform differently on tasks robust and sensitive to language variation.} 
Overall, tasks sensitive to language variation profit from tokenizers that encode more variation through 
a larger vocabulary size than 4k (\S\ref{sec:result_vocab}). %
Note that the best-performing tokenizer settings are not always consistent across the individual variation-robust and variation-sensitive tasks (e.g., vocabulary size for GLUE+typo and GLUE in Table \ref{fig:sig_robust}). We suspect that this is due to differences in the types of language variation present in the specific task datasets (e.g., character-level tokens seem to be more robust to spelling variation). Future work should investigate different types of language variation individually (e.g., spelling vs. lexical variation).

\begin{table*}[t!]
    \centering\small
        
        \begin{tabular}{ll|ccccc|c}
            \toprule
             & \textbf{Model} 
             & \textbf{AV (acc) $\uparrow$} 
             & \textbf{PAN (acc) $\uparrow$} 
             & \textbf{CORE (acc) $\uparrow$} 
             & \textbf{NUCLE (F1) $\uparrow$} 
             & \textbf{Dialect (F1) $\uparrow$} 
             & \textbf{AVG} \\ 
            \midrule
            \multirow{4}{*}{\textbf{Corpus}}  
            
              & PMed 
                & 81.7 $\pm$ 0.1 
                & 65.2 $\pm$ 0.5 
                & 55.9 $\pm$ 0.6 
                & 21.8 $\pm$ 1.2 
                & 87.9 $\pm$ 0.1 
                & 62.5 $\pm$ 0.4 \\
                
              & Wiki 
                & 81.9 $\pm$ 0.1 
                & 65.5 $\pm$ 0.5 
                & 55.5 $\pm$ 0.6 
                & \textbf{23.5} $\pm$ 0.2 
                & \textbf{88.9} $\pm$ 0.5 
                & 63.1 $\pm$ 0.2 \\
                
              & Twitter 
                & \textbf{82.9} $\pm$ 0.6 
                & \textbf{66.7} $\pm$ 0.4 
                & \textbf{56.5} $\pm$ 0.6 
                & 21.4 $\pm$ 0.9 
                & 88.3 $\pm$ 0.2 
                & \textbf{63.2} $\pm$ 0.2 \\

            \cmidrule{2-8}
            \multirow{5}{*}{\textbf{Pre-Tok.}}  
            
              & \fakesc{no}
                & 81.8 $\pm$ 0.3 
                & 59.9 $\pm$ 1.0 
                & 51.7 $\pm$ 0.2
                & 16.3 $\pm$ 0.2 
                & 77.3 $\pm$ 0.2 
                & 57.4 $\pm$ 0.2 \\
            
              & \fakesc{ws}
                & 75.5 $\pm$ 0.9 
                & 66.1 $\pm$ 0.4 
                & 55.1 $\pm$ 0.9 
                & \textbf{23.0} $\pm$ 0.1 
                & \textbf{88.8} $\pm$ 0.2 
                & 61.7 $\pm$ 0.3 \\
                
              & \fakesc{\_ws}
                & 82.5 $\pm$ 0.3 
                & 66.3 $\pm$ 1.6 
                & \textbf{56.6} $\pm$ 0.7 
                & 22.6 $\pm$ 0.5 
                & 88.4 $\pm$ 0.2
                & \textbf{63.3} $\pm$ 0.3 \\
              & \fakesc{llama3} 
                & {82.5} $\pm$ 0.2 
                & \textbf{66.9} $\pm$ 0.6 
                & 56.5 $\pm$ 0.6 
                & 21.1 $\pm$ 1.2 
                & 88.6 $\pm$ 0.0 
                & {63.1} $\pm$ 0.3 \\          
              & \fakesc{gpt2}
                & \textbf{82.6} $\pm$ 0.5 
                & 66.6 $\pm$ 1.2 
                & {56.3} $\pm$ 1.2 
                & {21.8} $\pm$ 0.9 
                & 88.4 $\pm$ 0.6 
                & 63.1 $\pm$ 0.4 \\
            \cmidrule{2-8}
            \multirow{5}{*}{\textbf{Size}}  
            
              & 500 
                & 78.2 $\pm$ 0.9 
                & 62.6 $\pm$ 0.2 
                & 51.1 $\pm$ 0.6 
                & 13.1 $\pm$ 0.9 
                & 85.6 $\pm$ 0.4 
                & 58.1 $\pm$ 0.6 \\
                
              & 4k 
                & 81.9 $\pm$ 0.1 
                & \textbf{67.8} $\pm$ 0.6 
                & 55.1 $\pm$ 1.0 
                & 17.4 $\pm$ 2.8 
                & 87.9 $\pm$ 0.7 
                & 62.0 $\pm$ 0.6 \\
                
              & 32k 
                & {82.6} $\pm$ 0.5 
                & 66.6 $\pm$ 1.2 
                & \textbf{56.3} $\pm$ 1.2  & {21.8} $\pm$ 0.9 
                & \textbf{88.4} $\pm$ 0.6 
                & \textbf{63.1} $\pm$ 0.4 \\
                
              & 64k 
                & \textbf{82.7} $\pm$ 0.4 
                & 67.2 $\pm$ 0.7 
                & 54.9 $\pm$ 1.5 
                & \textbf{22.0} $\pm$ 1.1 
                & {88.1} $\pm$ 0.6 
                & {63.0} $\pm$ 0.6 \\
                
              & 128k 
                & 80.1 $\pm$ 2.2 
                & 62.4 $\pm$ 3.4 
                & 51.2 $\pm$ 2.2 
                & 19.0 $\pm$ 2.7 
                & 81.5 $\pm$ 5.1 
                & 58.9 $\pm$ 3.0 \\
            
            \bottomrule
    \end{tabular}
    \caption{
        \textbf{Performance on tasks requiring sensitivity to language variation.} \label{tab:Style-performance}
        We display BERT performances on our Authorship Verification (AV), the PAN, the CORE, and the multi-Dialect dataset. We provide the mean and standard deviation ($\pm$) over three seeds respectively. %
        For the averaging column (AVG), italics indicate tokenizers with no statistically significant difference from the best-performing tokenizer (cf.~Figure \ref{fig:sig_robust}).
    }
\end{table*}

\subsection{Pre-Tokenizers} \label{result_pretokenizer}

\textbf{Pre-tokenizers have the greatest influence on performance.} For both task types, the range of performance values is largest for pre-tokenizers, second largest for vocabulary size and smallest for fitting corpus. This is surprising as, to the best of our knowledge, pre-tokenizers have received the least attention in previous work. 
For both task types, using pre-tokenizers improves performance over using no pre-tokenizer.

\textbf{
    Pre-tokenizer performance differs more between individual tasks than between task types. %
} 
For both task types, \fakesc{\_ws} is among the best performing pre-tokenizers.  Combining leading whitespaces with letters (\fakesc{\_ws}) generally improved performance over isolating whitespaces (\fakesc{ws}).  Notably, \_ws cuts the input tokens to a model in half as single whitespaces between words are not separate tokens anymore. 
For dialect and spelling-transformed GLUE, CORE, NUCLE and Dialect, the whitespace-based pre-tokenizers perform the best. For AV, PAN and the original GLUE tasks, \fakesc{gpt2} and \fakesc{Llama} perform better.
Tasks like grammatical error detection (NUCLE) could be seen as identifying deviations from a standard, and might benefit from pre-tokenizers that include tokens with typical combinations of Unicode character categories.
One of the main differences of \fakesc{gpt2} and \fakesc{Llama} to other pre-tokenizers is that the combination of different Unicode character categories (e.g., punctuation and letters) within the same token is less often allowed. By preventing combinations of categories like punctuation and letters, the vocabulary of \fakesc{gpt2} can include a broader range of tokens that only consist of letters, e.g.,  ``\_queer'' in Table \ref{tab:vocabulary-build}. This could explain the better performance of \fakesc{gpt2} on the original GLUE task. In contrast, \fakesc{llama3} also allows the mixing of one initial punctuation mark with letters, e.g., including ``'all'' in Table \ref{tab:vocabulary-build}, which after manual inspection seems to especially help \fakesc{llama3} solve PAN. Future work could investigate different individual tasks and the influence of pre-tokenizers individually.

\subsection{Fitting Corpus}

\textbf{On tasks requiring sensitivity to language variation, Twitter performs best.} This aligns with our expectation that the Twitter corpus includes more spelling variation and a larger set of language varieties than PubMed and Wikipedia. Interestingly, Wikipedia performs the best on NUCLE and Dialect. Grammatical error detection (NUCLE) could be seen as identifying deviations from a standard, and might benefit from training on corpora with few grammatical errors --- like Wikipedia.

\textbf{On tasks requiring robustness to language variation, Twitter performs surprisingly well.} Interestingly, Twitter performs indistinguishably from other fitting corpora on tasks that require robustness to language variation. Originally, we expected a more standardized corpus like Wikipedia to perform better on the original GLUE task, as it should lead to less spelling variation and thus more ``full words'' in the vocabulary (e.g., ``precursor'' in Table~\ref{tab:vocabulary-build}). %

\subsection{Vocabulary Size} \label{sec:result_vocab}

\textbf{A larger vocabulary size might be useful for tasks requiring sensitivity to language variation.} Performance peaked at 4k for tasks requiring robustness to language variation and at 32k for tasks requiring sensitivity to language variation. It seems that a smaller vocabulary size is sufficient to learn common standard forms and more tokens are needed to include style and form variations. For example, ``jumper'' is only included with vocabulary size 64k, see Table \ref{tab:vocabulary-build}. Note that optimal vocabulary sizes might scale with model size \cite{tao2024scaling}. %
For both task types, the standard deviation is highest for the smallest (500) and the largest vocabulary size (128k).

\textbf{Smaller vocabulary is more robust to spelling variation.} The vocabulary size of 500 consistently performs worse than larger vocabulary sizes. It is possible that BERT struggles
with long input sequences and 
learning to compose words for small vocabulary sizes. An exception is the GLUE+typo task, where the tokenizer of size 500 performs best. This is consistent with character-based tokenizers being more robust to spelling variation (cf.~\S\ref{sec:vocab-size}). %

\section{Pre-evaluating Downstream Tokenizer Impact} \label{sec:tokenizer-stats}

\begin{table}[t!]
    \centering\small
    \begin{tabular}{p{1cm}ll}
        \toprule
         \textbf{Group} & \textbf{Choice} & \textbf{Examples of tokens}\\
        \midrule
        \multirow{3}{*}{Corpus}
            & Wiki & 
                \_queer \_precursor \_l ma o
                \\ %
            & Twitter & %
                \_queer \_prec urs or \_lmao
            \\ %
            & PubMed & %
                \_qu e er \_precursor \_l ma o
            \\ %
        \midrule
        \multirow{4}{*}{Pre-Tok}
            & \fakesc{no} & %
                \_ y'all\_ que er\_ b 4 %
            \\  %
            & \fakesc{ws} & %
                \_ y'all \_ que er \_ b4 %
                \\
            & \fakesc{\_ws} & %
                \_y'all \_que er \_b 4 %
                \\ 
            & \fakesc{llama3} & %
                \_y 'all \_que er \_b 4 %
                \\
            & \fakesc{gpt2} & %
                \_y ' all \_queer \_b 4 %
                \\
        \midrule
        \multirow{4}{*}{Size}
            & 500 & %
                \_t e h \_l g b t q i a + \_j um p er
                \\  
            & 4k & %
                \_te h \_l g bt q ia + \_j um per
                \\
            & 32k & %
                \_te h \_l g bt q ia + \_j umper
                \\  
            & 64k & %
                \_teh \_lg bt q ia + \_jumper
                \\
            & 128k & %
                \_teh \_lgbt q ia + \_jumper
                \\
        \bottomrule
    \end{tabular}
    \caption{\textbf{Examples of vocabulary differences.} We display examples of sequences tokenized by tokenizers with varying fitting corpora, pre-tokenizers and vocabulary size. We represent whitespaces within tokens as \_. The domain of the fitting corpus affects the set of unique words in the vocabulary (e.g., ``queer'' is part of Twitter but not PubMed). The size determines the number of ``rarer'' words in the fitting corpus that are added to the vocabulary (e.g., the British variant ``jumper'' of ``sweater'' is only added with size 64k). The pre-tokenizer determines what types of words can be part of the vocabulary (e.g., ``b4'' can not be part of  \fakesc{gpt2} and \fakesc{llama3}). Note that allowing fewer Unicode character category combinations (e.g., numbers and letters) increases the broadness of the vocabulary %
    (e.g., in our fitting corpora, \fakesc{gpt2} 
    allows for fewer combinations than other pre-tokenizers and is the only one to include ``queer'' in its vocabulary).
    } \label{tab:vocabulary-build} 
\end{table}

A popular alternative to training several larger LMs to evaluate tokenizers is using fast \textit{intrinsic} measures on the token distribution on the downstream task corpus. %
However, common intrinsic measures are not consistently correlated with downstream performance \cite{rust-etal-2021-good,zouhar-etal-2023-tokenization,schmidt-etal-2024-tokenization}. 
Further, they do not make use of  task labels and can thus be considered \textit{task agnostic}. 
Imagine a corpus on which you want to perform both a task requiring robustness to language variation and a task requiring sensitivity to language variation (cf. Figure~\ref{fig:semantics_form}). Task-agnostic intrinsic measures like Corpus Token Count and Rényi efficiency \cite{zouhar-etal-2023-tokenization}  will always predict the same performance for both tasks. 
However, we found that the type of task can have an influence on what tokenizer settings work better (cf.~\S\ref{sec:result}). %

\subsection{Logistic Regression Measure}

We experiment with a task-dependent measure: the performance\footnote{We use the same F1 and accuracy performance metrics as for BERT.} of logistic regression with task labels as the dependent variable. Specifically, we use a tokenizer's vocabulary as the feature set with a bag-of-token approach. For tasks with two texts as inputs, token combinations between sentence pairs serve as features. For multi-class and multi-label tasks, we use one-vs-rest models. See Appendix~\ref{app:log-regression} for details.

\subsection{Results}

We display the Pearson correlation of Rényi efficiency, Corpus Token Count and logistic regression performances with the fine-tuned BERT performances in Table~\ref{tab:correlation}. Similar to \newcite{schmidt-etal-2024-tokenization} we find a light negative correlation of LM performances with Rényi efficiency on NLU tasks. 

\textbf{Corpus token count has a negative correlation for tasks requiring robustness and a positive correlation for tasks requiring sensitivity to language variation.} %
Corpus Token Count and correlated measures are often used to assess the compression quality of a tokenizer on a reference corpus or downstream task corpus. A higher value entails more tokens in the corpus and thus worse compression. Better compression is commonly believed to be a sign of a better tokenizer, and has often been thought to be correlated with better downstream performance \cite{rust-etal-2021-good,ali-etal-2024-tokenizer,velayuthan-sarveswaran-2025-egalitarian}. %
We show that the correlation is flipped for tasks that are robust and sensitive to language variation highlighting the difference in tokenizer requirements for the two task types. 

\textbf{RQ2: Logistic regression correlates with downstream performance.} Among the three considered measures, we find the highest correlation between logistic regression performances and BERT performances. Additionally, logistic regression is similarly correlated for both task types and can compare tokenizers of varying vocabulary sizes, which is not possible with Rényi efficiency and Corpus Token Count (for details refer to Appendix \ref{app:intrinsic_vocab-size}). Note that we by no means question the usefulness of measures like Corpus Token Count or Rényi efficiency for assessing different tokenizers and what they do. However, they might be less suited to estimate downstream performance without additional modifications.

\begin{table}[t!]
    \centering\small
    \begin{tabular}{l rrr}
        \toprule
            \textbf{Measure} & \textbf{Robust} & \textbf{Sensitive} \\
        \midrule
            Rényi Efficiency  & -.22 & -.03 \\ %
            Corpus Token Count  & -.45 & .37 & \\ %
            logistic regression & .85 & .84 \\ %
         \bottomrule
    \end{tabular}
    \caption{\textbf{Correlation of intrinsic measures with BERT performances.} Correlations between the Rényi Efficency ($\alpha=2.5$), Corpus Token Count, %
    and logistic regression predictions with the performance of the finetuned BERT models on the tasks robust to language variation and the tasks sensitive to language variation respectively. %
    Logistic regression has the highest correlation values across all tasks. A higher Corpus Token Count is negatively correlated with performance on tasks robust to language variation and positively correlated with tasks sensitive to language variation.
    }
    \label{tab:correlation}
\end{table}

\section{Conclusion}

In this work, we investigated tokenizer settings for tasks that require robustness and tasks that require sensitivity to language variation.
BPE settings perform differently on the two task types. 
We make three practical suggestions for selecting tokenizer settings: (1) Pay the most attention to the pre-tokenizer. It influences how Unicode character categories can be combined (e.g., ``b4'' in Table~\ref{tab:vocabulary-build}), %
and how many different words are ultimately part of the vocabulary (e.g., ``\_queer'' in Table~\ref{tab:vocabulary-build}). (2) Choose a bigger vocabulary size for settings that require sensitivity to language variation. (3) Use a small machine learning classifier --- e.g., a logistic regression classifier --- to evaluate how different tokenizers affect performance on tasks in your domain. For example, for general-purpose language models, pick a tokenizer that this model consistently predicts to perform well across both variation-robust and variation-sensitive tasks.
The best tokenizer settings might be sensitive to what types of language variation (e.g., lexical vs. syntactic) are present. We think it is crucial to investigate different types of language variation individually in future work to ultimately build better, fairer and more robust LLMs.

\section*{Limitations}

Note that our taxonomy of tasks relies on splitting tasks into semantic-focused tasks (i.e., considering what is said) and form-based tasks (i.e., considering how it is said). However, a strict distinction is challenging since most tasks are best solved using a mixture of typical content and typical form signals. For example, recognizing the lengthening of words (e.g., ``cooool'') can be helpful for the semantic task of sentiment classification \cite{brody-diakopoulos-2011-cooooooooooooooollllllllllllll} and content information (e.g., the topic) can improve authorship verification performance \cite{wegmann-etal-2022-author}. Still, this distinction clarifies the main signals that our different task types should rely on and enables us to compare tokenizer settings on these two different types of tasks. 

We evaluate tokenizers on 110M parameter encoder models, because they tend to reach higher performance for classification tasks in low parameter settings. However, we risk not accounting for emergent capabilities of popular larger generative language models. For example, the number of context tokens for recent models are generally much larger than the original 512 tokens of BERT base. %

We find that logistic regression correlates with downstream performance on classification tasks. For more complex tasks like text generation, logistic regression probably will not be applicable. However, we still see it as a great benefit that logistic regression can enable us to quickly test tokenizers on a variety of different tasks and make a more informed decision on what tokenizers to test more rigorously. %

For tasks requiring robustness to language variation, many more tasks could have been included. 
For example, future work could include challenge NLI datasets, e.g., ANLI \cite{nie-etal-2020-adversarial}. Further, future work could investigate different types of tasks such as question answering or arithmetic tasks.  %
For tasks requiring sensitivity to language variation, we originally planned to include more classification tasks that are tied to well-established factors of language variation, e.g., age prediction and geographic location prediction \cite{nguyen-etal-2016-survey}. However, we repeatedly encountered difficulties with cleanly separating semantic from form-based cues. For age prediction and location prediction specifically, logistic regression models made extensive use of content words. Based on the coefficients of our logistic regression models, we excluded tasks that we expected to mainly require sensitivity to language variation, but where the models often relied on content words. %

We varied one tokenizer setting at a time while leaving the other two on their ``default values'' (i.e., miscellaneous fitting corpus, \textsc{gpt2} pre-tokenizer and vocabulary size of 32k), in order to isolate the effect of each individual setting without exhaustive testing of all possible combinations. Future work could explore interdependencies between tokenizer settings thorugh more comprehensive testing. 

The tokenizer settings seem to have different effects when the pre-training corpus and fitting corpus differ. %
Specifically, the pre-training corpus might influence in how far a model can leverage the diversity encoded in the tokenizer. 
See Appendix \ref{app:webbook-pretraining} for further experiments. Note, however, that recent work suggests to only use vocabulary that appears in the training corpus \cite{land-bartolo-2024-fishing} -- which is unlikely when different corpora are used for model training and tokenizer fitting. 

\section*{Ethical Considerations}
Our pre-training and fitting corpora are largely based on publicly shared and accessible datasets from popular online forums and web pages (e.g., Wikipedia, Reddit, Common Crawl, ...). Unfortunately, these datasets were mostly collected without explicit consent from users and might lead to (among others) privacy concerns. Individuals might be identifiable from their written texts. However, we hope that the risks of reusing already published datasets are minimal. We collected tweets from Twitter in 2021 using Twitter's official API, but opt to not share them publicly.
While we aim to include different language varieties in our datasets, they might not be representative of English language use across different social groups. For example, we expect a skew towards straight, white, American, young and male authors. We caution against using our datasets and tasks for general claims about broad selection of language varieties. 
We confirm to have read and that we abide by the ACL Code of Ethics. Beside the mentioned ethical considerations, we do not foresee immediate risks
of our work.

\section*{Acknowledgements}  
    We thank the anonymous ARR reviewers for their constructive comments. 
    We thank the colleagues in the various NLP Groups at Utrecht University and the Blablablab at the University of Michigan and, specifically,  %
    Kees van Deemter, Antal van den Bosch, Yupei Du, Qixiang Fang, Melody Sepahpour-Fard, Shane Kaszefski Yaschuk, Elize Herrewijnen, Sanne Hoeken, Hugh Mee Wong, Jian Zhu and Pablo Mosteiro for, among others, feedback on writing and presentation. %
    We thank colleagues at EMNLP 2024 for their insights into tokenization, and, specifically, Vilém Zouhar. 
    We thank Laura Wegmann for feedback on writing.
    This research is supported by the ``Digital Society - The Informed Citizen'' research programme, which is (partly) financed by the Dutch Research Council (NWO), project 410.19.007. %
    This research is supported in part by the Office of the Director of National Intelligence (ODNI), Intelligence Advanced Research Projects Activity (IARPA), via the HIATUS Program contract \#2022-22072200006. The views and conclusions contained herein are those of the authors and should not be interpreted as necessarily representing the official policies, either expressed or implied, of ODNI, IARPA, or the U.S. Government. The U.S. Government is authorized to reproduce and distribute reprints for governmental purposes notwithstanding any copyright annotation therein.
\bibliography{custom, anthology}

\appendix

\clearpage\newpage

\begin{table}[t!]
    \centering\small
    \begin{tabular}{p{1cm}lp{3.8cm}}
        \toprule
         \textbf{Group} & \textbf{Choice} & \textbf{Examples of tokens}\\
        \midrule
        \multirow{4}{*}{Corpus}
            & Wikipedia & ebruary \_Retrieved  arliament\\ %
            & Twitter & \_[joy] [loudly-crying-face][loudly-crying-face] [heart] BTS\\ %
            & PubMed & \_effec \_lymphadenopathy\\ %
            & Misc. & %
                \textbackslash t\textbackslash t \ \textbackslash r \textbackslash n  \_differe \\
        \midrule
        \multirow{4}{*}{Pre-Tok}
            & no & \_that\_ at\_ in\_this\_case\_ \\  %
            & ws & you're that's took develo \\
            & \_ws & \textbackslash nI \_I'm \textbackslash nWhat \_devices.\\ 
            & \fakesc{gpt2} & \_..., ensional >"; \_127 \\
            & \fakesc{llama3} & 433 \_\{\textbackslash n \_*\textbackslash n .apache  \\
        \midrule
        \multirow{4}{*}{Size}
            & 500 & ! \$ 0 1 2 A B C is he in \\  
            & 4k & age very \_will \_would \\
            & 32k & \_surveillance \_Vietnam \\  
            & 64k & CAN 322 \_infuri \\
            & 128k & motherboard \_narcotics \\
        \bottomrule
    \end{tabular}
    \caption{\textbf{Examples of unique tokens for each tokenizer choice.} The displayed tokens are unique to the given tokenizer in their respective group, except for the different vocabulary sizes. The tokens of the smaller vocabulary size is always included in the bigger vocabulary size. %
    We represent whitespaces within tokens as \_. We represent emojis within [] with their textual descriptions.
    } \label{tab:tokenizer-choices} 
\end{table}

\section{Tokenizer} \label{app:tokenizer}

In this section, we provide additional information on the tokenizer settings we investigate. See Table~\ref{tab:tokenizer-choices} for examples of unique tokens for each setting.

\subsection{Fitting Corpora} \label{app:fitting-corpora} \label{app:mixed-dist}

See an overview of the fitting dataset sizes in Table~\ref{app:tab-fitting-corpora}. Usually, the fitting corpus for the tokenizer and the training corpus for the language model with that tokenizer are the same. As a result the size of the fitting corpus often varies as widely as the size of training datasets. We aim for 1.5 billion tokens for all fitting corpora. The MiniPile \cite{kaddour2023minipile} dataset used for fitting in \newcite{schmidt-etal-2024-tokenization} is of similar magnitude. %
We further display text examples for each dataset in Table \ref{tab:data-examples}. Dataset sizes vary in less than 1\% of word count. The variance in word count is an artifact of dataset creation from several documents with lenient word count limits.%

\begin{table}[t]
    \centering
    \small
    \begin{tabular}{l r r r}
        \toprule
        \textbf{Source} & \textbf{Train} & \textbf{Dev} & \textbf{Test} \\
        \midrule
        Wikipedia & 1,469,999,792 & 15,000,029 & 15,000,087 \\
        Twitter & 1,470,004,662 & 15,000,048 & 15,000,057 \\
        PubMed & 1,469,999,499 & 15,000,106 & 15,000,501 \\
        Mixed & 1,477,872,323 & 15,080,505 & 15,080,919 \\
        \bottomrule
    \end{tabular}
    \caption{\textbf{Fitting corpora with similar word counts.} We compare three fitting corpora for tokenizers. Word count is calculated using white-space splitting. The size of the fitting corpora are not exactly the same when it comes to word count. But variations in word count are below 1\% and should not affect the vocabulary of the tokenizer fitted on them.}
    \label{app:tab-fitting-corpora}
\end{table}

\begin{table*}[h]
    \centering\scriptsize
    \begin{tabular}{lp{8cm}rl}
    \toprule
    \textbf{Source} & \textbf{Text} & \textbf{word count} & \textbf{domain} \\
    \midrule
    \multirow{1}{*}{Wiki} 
         &Mary Jane Christie Serrano (c. 1840 – 1923) was a writer, poet and considered ... & 24 & - \\ %
    \midrule
    \multirow{1}{*}{Twitter} 
         & Where are the top places in Broward or Palm Beach? [thinking-face][eyes] & 10 & - \\ %
    \midrule
    \multirow{1}{*}{PMed} 
         & ... Myoelectrical activity of the gut has been studied in the postoperative period ... %
         & 132 & -\\
    \midrule
    \multirow{18}{*}{Misc.} 
       & Israel, as usual, wants American forces to fight a bloody war against Iran. ... & 66 & nytimes\\
       & \&gt;In Israel, my grandfather fought for its life. The people down the street fought ...  & 581 & reddit \\ %
        & Q:\textbackslash n\textbackslash nHow can I determine the current focused process name and version in C\#... & 122 & StackEx. \\
        & I read the audio version of this story and loved it. ... & 53 & goodreads \\
        & ... always get 100 test cases (or whatever the default number of test is)?\textbackslash n\textbackslash nJanek ... & 134 & gmane \\
        & there is Sydney waiting to enter. Their eyes meet. “Maggie! I was just ... & 998 & ao3 \\
        & ... Abstract. How many n-orthants can be intersected in the n-dimensional ... & 2048 & s2orc \\
        & from torch import optim as optim\textbackslash n \textbackslash n from geoopt.optim.mixin import  ... & 115 & GitHub \\
        & if you're looking for enhancement cores in the game one really useful way  ... & 102 & YouTube\\
        & I have a lot of ties. But my favorite one. My favorite tie is owned by  ... & 183 & blogcorpus\\
        & Witnesses interviewed: 3 (N°1141, 1142, 1143). Nikolay S.( N°1142): "After the ... & 72 & Pile-CC\\
        & It came in perfect condition and it is very soft. & 9 & amazon \\
        & I'm sure that the procurement people are doing the best job they can ... & 94 & sfu-socc \\
        & ... Sort -11, -1, 0, -3, 5 in decreasing order ... & 1645 & DM Maths \\
        & ... "But the killer isn't the Russian army." "It's the subzero temperatures." ... & 1978 & OpenSubtitles \\
        & ... THE\textbackslash n\textbackslash n LIFE\textbackslash n\textbackslash n OF\textbackslash n\textbackslash n GEORGE WASHINGTON,\textbackslash n\textbackslash n COMMANDER IN ... & 2048 & Gutenberg \\
        & Rumson, NJ – December 2013 What started in 2003 as a group of mostly Christian ... & 592 & realnews \\ %
    \bottomrule
    \end{tabular}
    \caption{\textbf{Dataset examples.}\label{tab:data-examples} We show text examples for all used fitting corpora. %
    }
\end{table*}

\begin{table*}[h]
    \centering
    \scriptsize
    \begin{tabular}{l l r r r }
        \toprule
        \textbf{Genre} & \textbf{Domain} & \textbf{Train} & \textbf{Dev} & \textbf{Test}\\
        \midrule
        Forum & Reddit & 245M %
            & 2.5M %
            & 2.5M %
            \\
        Literature & AO3 & 147M %
            & 1.5M %
            & 1.5M %
            \\
        Literature & Gutenberg before 1919 
            & 49M %
            & 0.5M %
            & 0.5M %
            \\
        News & Realnews & 147M %
            & 1.5M %
            & 1.5M %
            \\
        News/Comments & NYTimes \& Comments 
            & 24M %
            & 0.3M %
            & 0.3M %
            \\
        News/Comments & SFU-SOCC 
            & 3M %
            & 0.03M %
            & 0.02M %
            \\
        Q\&A & StackExchange 
            & 196M %
            & 2.0M %
            & 2.0M %
            \\
        Reviews & Goodreads 
            & 49M %
            & 0.5M %
            & 0.5M %
            \\
        Reviews & Amazon 
            & 49M %
            & 0.5M %
            & 0.5M %
            \\
        Mails & Gmane & 147M %
            & 1.5M %
            & 1.5M %
            \\
        Transcripts & YouTubeCommons & 98M %
            & 0.9M %
            & 1.0M %
            \\
        Transcripts & OpenSubtitles & 49M %
            & 0.5M %
            & 0.5M %
            \\
        Code & GitHub 
            & 49M %
            & 0.5M %
            & 0.5M %
            \\
        Science & S2ORC 
            & 98M %
            & 1.0M %
            & 0.9M %
            \\
        Blogs 
            & BlogCorpus & 10M %
            & 0.1M %
            & 0.1M %
            \\
        Raw Text Webpages & CommonCrawl & 98M %
            & 1.0M %
            & 1.0M %
            \\
        Mathematics & DM Mathematics & 20M %
            & 0.2M %
            & 0.2M %
            \\
        \midrule
        \multicolumn{2}{r}{\textbf{Total:}} & \textbf{1,478M} %
            & \textbf{15.1M} %
            & \textbf{15.1M} %
            \\
        \bottomrule
    \end{tabular}
    \caption{\textbf{Miscellaneous dataset statistics.}}
    \label{tab:updated_dataset_information_corrected}
\end{table*}

\textbf{Miscellaneous.} Miscellaneous consists of Reddit \cite{baumgartner2020pushshift}, %
literature sources (fanfictions from ao3\footnote{fanfictions until 2019 from Archive of Our Own \url{https://archiveofourown.org/}, downloaded from \url{https://archive.org/download/AO3_story_dump_continuing} in 2023, filtered for English language using AO3 tags. Dataset was removed but should be re-creatble using tools like \url{https://github.com/nianeyna/ao3downloader}.
}, \citealp{gao2020pile}'s books before 1919 from the Gutenberg project), news articles and comments (\citealp{zellers2019grover}'s realnews, 2020 NYTimes articles and comments\footnote{\url{https://www.kaggle.com/datasets/benjaminawd/new-york-times-articles-comments-2020} , minimum length filter of 250 was applied
}, \citealp{kolhatkar2020sfu}'s sfu-socc), question answering (\citealp{gao2020pile}'s StackExchange), reviews (\citealp{hou2024bridging}'s Amazon and \citealp{wan2018item}'s GoodReads %
reviews), mails (\citealp{bevendorff-etal-2020-crawling}'s Gmane%
), transcripts (YouTubeCommons\footnote{\url{https://huggingface.co/datasets/PleIAs/YouTube-Commons}} and \citealp{gao2020pile}'s OpenSubtitles), blogs (\citealp{schler2006age}'s blogcorpus%
), raw text from webpages (\citealp{gao2020pile}'s Common Crawl), science articles (\citealp{lo-etal-2020-s2orc}'s s2orc), code and mathematics (\citealp{gao2020pile}'s GitHub and DeepMind Mathematics). See the share of different domains in Table \ref{tab:updated_dataset_information_corrected}. %

\subsection{Pre-Tokenizer} \label{app:tokenizer-pretokenizer}
    See the regular expressions defining the different considered pre-tokenizers in Table \ref{app:pre-tokenizer}. Differences affect mostly whitespace, contraction, punctuation and number handling. 

    \begin{table*}[h!]
        \centering
        \scriptsize
        \begin{tabular}{l l p{9cm}}
            \toprule
            \textbf{name} & \textbf{RegEx} & \textbf{Example Text}\\
            \midrule
            \fakesc{no} & -
                & well...\_\$3000\_for\_a\_tokenizer\_isn$^\backprime$t\_cheapz\_\#lol\_:)\textbackslash n\textbackslash nhttps://en.wikipedia.org/wiki/Sarcasm \\
            \midrule
            \fakesc{ws} & {\begin{lstlisting}
    \s+\end{lstlisting}} 
                &  well... \_ \$3000 \_ for \_ a \_ tokenizer \_ isn$^\backprime$t \_ cheapz \_ \#lol \_ :)\textbackslash n\textbackslash nhttps://en.wikipedia.org/wiki/Sarcasm  \\
            \midrule
            \fakesc{\_ws} & {\begin{lstlisting}
    \s+(?!\S)|\s+\end{lstlisting}} 
                & well... \_\$3000 \_for \_a \_tokenizer \_isn$^\backprime$t \_cheapz \_\#lol \_:) \textbackslash n \textbackslash nhttps://en.wikipedia.org/wiki/Sarcasm  \\
            \midrule
            \fakesc{gpt2} & \vspace{-1.3\baselineskip}{\begin{lstlisting}
        's|'t|'re|'ve|'m|'ll|'d 
    | ?\p{L}+| ?\p{N}+ 
    | ?[^\s\p{L}\p{N}]+
               |\s+(?!\S)|\s+\end{lstlisting}} 
    & well ... \_\$ 3000 \_for \_a \_tokenizer \_isn $^\backprime$ t \_cheapz \_\# lol \_:) \textbackslash n \textbackslash n https :// en . wikipedia . org / wiki / Sarcasm \\
            \midrule
            \fakesc{llama3} & \vspace{-1.4\baselineskip}\begin{lstlisting}
    (?i:'s|'t|'re|'ve|'m|'ll|'d)
    |[^\r\n\p{L}\p{N}]?\p{L}+
    |\p{N}{1,3}
    | ?[^\s\p{L}\p{N}]+[\r\n]*
    |\s*[\r\n]+|\s+(?!\S)|\s+\end{lstlisting}  
    & well ... \_\$ 300 0 \_for \_a \_tokenizer \_isn $^\backprime$t \_cheapz \_\# lol \_:)\textbackslash n\textbackslash n https :// en .wikipedia .org /wiki /Sarcasm  \\
            \bottomrule
        \end{tabular}
        \caption{\textbf{Investigated Pre-tokenizers.}\label{app:pre-tokenizer} The pre-tokenizers we investigate can be described with regular expressions. 
        We investigate using no pre-tokenizer (no), isolating whitespaces (ws), split on whitespaces including single leading whitespaces in non-whitespace tokens (\_ws), the pre-tokenizer used by GPT-2 (gpt2) and the pre-tokenizer used by Llama 3 (llama3). GPT-2 and Llama 3 mainly differ in contraction, URL, whitespace and number handling.
        We display how the investigated pre-tokenizer split an example text. We replace whitespaces with \_ to highlight pre-token boundaries with whitespace. 
        }
    \end{table*}

\twocolumn
\clearpage
\newpage 

\begin{table*}[th!]
    \centering\scriptsize
    \begin{tabular}{p{1cm}p{1.5cm}p{2cm}p{4cm}p{2.5cm}p{2cm}}
    \toprule
    \textbf{Name} & \textbf{Task} & \textbf{Source} & \textbf{Text 1} & \textbf{Text 2} & \textbf{label} \\
    \midrule
    \multirow{3}{*}{CORE} 
         & Register Classification
         & \cite{laippala2023register}
         & [...] %
         What do you do when you just cant seem to mix something? Hi, You have mixed it...you just didn't know when to stop and move on! [...] & - & Interactive Discussion \\ \midrule
    \multirow{3}{*}{
        AV
    }
        & Authorship Verification
        & our contribution
        & Hi,\textbackslash n\textbackslash nI am currently evaluating OTRS for use as a Helpdesk System.\textbackslash n\textbackslash nHowever I am a little confused about how best to set it up. [...] &  Hi,\textbackslash n\textbackslash nI am setting up Mailscanner not to deliver any spam to my users, but to\textbackslash nsend a spam report once a day %
        [...] 
        & Same Author \\ \midrule
    \multirow{4}{*}{PAN}
        & Author Change
        & \cite{ayele2024overview}
        & I'm not gonna watch the video. I gotta keep my sanity. With that being said, what could we call for that they haven't done? The cops have been fired and charged with murder. [...] 
        &  Maybe police agencies should be not federalized but under one agency. No more sherriffs no more local police. Just state police. [...] %
        & Author Change \\ \midrule
    \multirow{2}{*}{NUCLE}
        & Error Classifications
        & \cite{dahlmeier-etal-2013-building}
        & Chernobyl accident, happened in 1986, was a nuclear reactor accident. & - & [ArtOrDet, Trans]
        \\ \midrule
    \multirow{1}{*}{Dialect}
        & Dialect Classification
        & transformations with \cite{ziems-etal-2023-multi}
        & What the best things to do in Hong Kong one? & - & CollSgE 
        \\
    \bottomrule
    \end{tabular}
    \caption{\textbf{Tasks Sensitive to Language Variation.} \label{tab:sensitive} For each task, we show an example.}
\end{table*}

\section{Evaluation Tasks}
\subsection{Tasks Robust To Language Variation} \label{App:NLU}
\textbf{GLUE task selection.} We originally planned to pre-train BERT models and test them on the same GLUE tasks \cite{wang-etal-2018-glue} as the ones used for the original BERT model \cite{devlin-etal-2019-bert}, %
i.e., CoLA, SST-2, MRPC, STS-B, QQP, MNLI, QNLI and RTE. 
We removed CoLA \cite{CoLA} as it is the task of classifying linguistic acceptability. Models have to classify morphological, syntactic and semantic ``violations'' and the task can thus be expected to be sensitive to language variation. Further, we removed STS-B \cite{cer-etal-2017-semeval}. STS-B is a regression task for the semantic similarity between two sentences. However, we aimed to focus on classification tasks that could be modeled with logistic regression. 
After pre-training and fine-tuning on the remaining GLUE tasks, we also removed MRPC \cite{dolan-brockett-2005-automatically} and RTE from the evaluated tasks. %
We removed RTE because the standard deviation of $.04$ made the differences in performance for models %
unclear, the variation could be related to the small training dataset of only 2.5k instances. We further removed MRPC as it showed almost no variation between different tokenizers -- not even for the 500 vocabulary size. %
All remaining tasks had a train set size of at least 67k. %
Note that SST-2 was released in a parsed format %
resulting in a lowercased and pre-tokenized text which might affect results.

\textbf{GLUE+dialect.} We transform GLUE using  \texttt{Multi-VALUE} \cite{ziems-etal-2023-multi}.  \texttt{Multi-VALUE} has strong requirements on text formatting. When \texttt{Multi-VALUE} perturbations fail we leave the GLUE text in that row as is. Depending on the task this concerns between 3\% and 18\% of instances.

\subsection{Tasks Requiring Sensitivity To Language Variation} \label{app:sadiri}

See an example for each of the tasks requiring sensitivity to language variation in Table \ref{tab:sensitive}.

\textbf{Authorship Verification.} %
Similar to the Miscellaneous corpus \ref{app:fitting-corpora}, the authorship verification corpus consists of datasets taken from Amazon \cite{hou2024bridging}, AO3, GMane \cite{bevendorff-etal-2020-crawling}, 2020 NYTimes articles and comments, realnews \cite{zellers2019grover}, Reddit \cite{baumgartner2020pushshift}, StackExchange \citep{gao2020pile} and Wikipedia articles. Additionally, the corpus includes texts from Bookcorpus \cite{books1} and PubMed \cite{gao2020pile}. It totals about 40.8k train pairs, %
2.5k dev pairs %
and 4.8 test pairs. %
For the contrastive training task, we use the Supervised Contrastive Loss\footnote{
    SupConLoss as implemented in \texttt{pytorch\_metric\_learning}, see \url{https://kevinmusgrave.github.io/pytorch-metric-learning/losses/}
} 
\cite{khosla2020supervised} with a siamese setup, a batch size of 128, and a learning rate of 0.00001. We find the threshold best separating same author and distinct author pairs on the dev set and report accuracy on the test set. We release the dataset on Hugging Face.\footnote{Future HuggingFace Link.}

\textbf{PAN.} The dataset was extracted from Reddit and preprocessed by removing citations, markdown, emojis, hyperlinks, multiple line breaks and extra whitespace \cite{ayele2024overview}. Compared to the Authorship Verification task (where the classifiers may learn to rely on content cues, apostrophe encodings, whitespace encoding, etc., cf.~\citealp{wegmann-etal-2022-author}), this may be more difficult. The dataset was downloaded from \url{https://pan.webis.de/clef24/pan24-web/style-change-detection.html}.

\textbf{NUCLE.} The dataset was downloaded from {\url{https://www.comp.nus.edu.sg/~nlp/corpora.html}}.

\textbf{CORE.} We use 8 main register labels for multi-class prediction. We display an example for all considered CORE \cite{laippala2023register} labels in Table~\ref{app:CORE-examples}. The original CORE consists of main as well as sub-labels that make up a total of 56 labels in \newcite{laippala2023register}. However using all 56 in a multi-label setup proved too difficult for our BERT as well as the logistic regression models without further hyperparameter tuning. We decided for a multi-classifcation setup, limiting ourselves to  8 out of 9 main register labels, specifically we excluded the ``OTHER'' catgory. %
This reduced the train dataset from about 34k instances to 30k.
We split up texts of length > 250  to chunks of a maximum of 250 using whitespace splitting. Then, we perform stratified sampling with replacement for each class to additionally upweigh small classes. Note that this results in duplicates for the small classes up to a maximum of 10 occurrences of the same text.

\begin{table}[t!]
    \centering\scriptsize
    \begin{tabular}{p{5cm}p{1.5cm}}
    \toprule
    \textbf{Text} & \textbf{label} \\
    \midrule
          ... Sometimes, people just don't feel well. But if you don't feel well more than sometimes, it may be helpful to talk to someone about it.  ... & opinion \\ 
          \midrule
          ... A transportation advocacy group is circulating a list of 100 questions aimed at broadening the British Columbia government's consultation on coastal ferry services. ... & narrative \\
          \midrule
          ... I'm sure many people have hit this brick wall. What do you do when you just cant seem to mix something? Hi, You have mixed it...you just didn't know when to stop and move on! ... & interactive discussion \\
          \midrule
          'Always think of home': an introduction to the Buenos Ayres Notebook The Buenos Ayres Notebook takes its name from the city Buenos Aires ('Good Air' or 'Fair Winds'), ... & informational description/explanation \\
          \midrule
          ... It would be a pleasure just to know just a little bit moreoh oh I could grow quite fond of your acquaintance ... & lyrical \\
          \midrule
          An Acadian-style cabin constructed completely of rough sawed Southern Yellow Pine, surrounded by split rail fence.  ... & informational persuasion \\
          \midrule
          ... it can be easy or even enjoyable. Here is a guide on how to give an oral presentation in front of your class. Decide on a topic ...  & how-to/instructional \\
          \midrule
           ...  Kareem Ettouney, the art director at Media Molecule always said, Mash up, not mish mash! -------- -------- What kinds of challenges did you face with replicating LBP's iconic 2D puppet aesthetic into a 3D space? ...   & spoken \\
    \bottomrule
    \end{tabular}
    \caption{\textbf{Examples for main CORE labels.} \label{app:CORE-examples}
       We focus on the 8 main CORE \cite{laippala2023register} labels.
    }
\end{table}

\clearpage 
\newpage

\section{Modeling} \label{app:modeling}

\begin{table}[t!]
    \centering
    \small
    \begin{tabular}{rrrlrr}
        \toprule
        \textbf{GPU h} & \textbf{P} & \textbf{T} &  \textbf{loss} %
        & \textbf{steps} & \textbf{batch}  \\
        \midrule
         0h & 4.6M & 10M & 7.7 & 600 & 32\\ %
        1h & 42M & 100M & 6.5 & 6k & 32\\ %
        4h & 110M & 250M & 6.2 & 15k & 32 \\ %
        9.0 & 110M & 750M & 3.9 & 11k & 128 \\ %
        11.0 & 110M & 750M & \textbf{2.7} & 45k & 32 \\ %
         {12.3} %
             & 4.6M & 330M 
             &  {4.1}%
             & 80k & 256 \\ %
        {13.2} &110M & 750M &  3.2 & 14k & 128 \\ %
        13.6 & 4.6M & 3300M & 4.1 & 75k & 256 \\%  & 62.6 
        14.8%
            & 4.6M & 3300M & {4.1}%
            & 80k & 256 \\
        22.0 & 11.6M & 3300M & 3.2 & 75k & 256 \\ %

        \bottomrule
    \end{tabular}
    \caption{\textbf{Hyperparameters for BERT Pre-Training.} \label{tab:chinchilla-BERT} We compare the evaluation loss, %
    and GPU hours while varying the number of parameters, tokens, steps and batch size. %
    For similar GPU hours (between 11h-15h), using more pre-training tokens does not seem to improve performance as much as increasing model size. 
    Balancing the number of model parameters and tokens, as well as the number of steps seems crucial.
    }
\end{table}

\textbf{Compute Optimal BERT.}  %
We originally evaluated tokenizers on tiny BERT models, using 80k steps on a training dataset with 3.3B tokens during pre-training (second to last row in Table \ref{tab:chinchilla-BERT}). This corresponds to more than three epochs on the relatively large training dataset for a tiny BERT model with only 4.6M parameters. Using this setup we found that tokenizers with the largest vocabulary sizes repeatedly outperformed all other settings. For tiny BERT, models with larger vocabulary sizes also need to use orders of magnitude more parameters because of the larger embedding matrix. For tiny BERT, the model size rises to 17M for a vocabulary size of 128k. Were we using a non-optimal ratio between number of tokens and parameters for our invested 15 GPU hours? %
We experimented with different BERT model sizes P (tiny -- 4.6M, small -- 11.6M, base -- 110M), %
number of tokens T, batch sizes and number of steps. See the results in Table~\ref{tab:chinchilla-BERT}. 
We evaluate performance with the eval loss  %
on the held out set of the pretraining corpus consisting equally of Bookcorpus3 \cite{books1} and OpenWebText2 \cite{gao2020pile}. We use one tokenizer fit on the Miscellaneous corpus, using the gpt-2 pre-tokenizer and a vocabulary size of 32k tokens. For further hyperparameter choices, see the {Hyperparameters} paragraph.
We make the following observations: 
    For similar GPU hours (between 11h-15h), using exponentially more training corpus tokens T than parameters P for tiny BERT does not improve performance as much as increasing model size P. This and other model results might hint at Chinchilla's scaling law, that is, optimal token count scaling with the parameter size of a model for a fixed compute, specifically $T_{OPT}\approx P^{27/23}$ \cite{chinchilla}, also holding for smaller encoder models. %
    Further, for our low resource setting, using more steps, and thus weight updates, seems to be more important than using a large batch size. 
These conclusions seems promising but very tentative. Exhaustive pretraining experiments were out of scope for this work. Nevertheless, we think that finding compute-optimal settings to train small transformer models is crucial to efficiently evaluate tokenizers going forward.

\begin{table}[t!]
    \centering
    \small
    \begin{tabular}{p{1.5cm} r p{3.1cm}}
        \toprule
        \textbf{Source} & \textbf{Word Count} & \textbf{Excerpt}\\
        \midrule
        Book\-Corpus & 1,687,724,544 & ... visit and they all swore on a second blood oath that it wasn't them.\textbackslash n\textbackslash n``What about Phantom?'' Hell Girl asked. ... \\
        Open\-Web\-Text2 & 1,650,000,384 & ... The future of SA-affiliated club sports, a cappella, and Greek groups is uncertain after the All-Campus Judicial Council ruled Friday ... \\
        \midrule
        Total & 3,337,724,928 &  \\
        \bottomrule
    \end{tabular}
    \caption{\textbf{Pretraining Corpora.} Word count is calculated using white-space splitting.}
    \label{tab:pretrain-corpus}
\end{table}

\textbf{WebBook Corpus.} Our training consists of equal parts Bookcorpus3\footnote{\url{https://twitter.com/theshawwn/status/1320282149329784833}} \cite{books1} and OpenWebText2 \cite{gao2020pile}. See the statistics in Table \ref{tab:pretrain-corpus}. We randomly sample sequences of 512 words, %
totaling 3.3 billion words as in the original BERT paper \cite{devlin-etal-2019-bert}. We ensure English excerpts by removing books and web text that are not predicted as English language by using \texttt{langdetect}.\footnote{\url{https://github.com/Mimino666/langdetect}} %
We split of 2\% of the sampled data for held out dev and test sets to evaluate BERT pretraining.

\label{app:pretraining}
\textbf{Hyperparameters.} We keep the following pre-training settings the same as {BERT} \cite{devlin-etal-2019-bert%
}: Adam with learning rate of 1e-4, $\beta1$ = 0.9, $\beta2$ = 0.999, L2 weight decay of 0.01, learning rate warmup over the first 1\% of steps, linear decay of the learning rate, dropout probability of 0.1 on all layers.
However, we do not use the next sentence prediction objective and only train on MLM with mlm probability at 15\% as NSP proved to be inferior to MLM in later models \cite{liu2019roberta}. 
Originally, pre-training was performed on a set of 3.3 billion words over 40 epochs, we experiment with different number of tokens and steps in \S\ref{sec:pretrain}.
We use the architecture of the tiny, small, medium and base BERT model %
\citep{turc2019well} which consists of 4.6, 11.6, 42, and 110 million parameters respectively.\footnote{
    This might affect performances for models with especially big vocabulary sizes.
}

\section{Intrinsic Evaluation}

\label{app:log-regression} \textbf{Logistic regression.} We tokenize the task texts. Then, we use the resulting tokens as features for the logistic regression. We use a bag-of-tokens approach and do not consider word order. We do not consider frequency but use binary information on token presence. We use Cartesian products of tokens for NLI tasks. Specifically, given two sentences in an NLI setting, we generate all possible combinations of individual tokens from both sentences and use these as features. For PAN and Authorship Verification, we limit the Cartesian product to tokens that appear in both sentences and tokens that only appear in one sentence.\footnote{PAN and AV had considerably longer texts than the GLUE NLI tasks leading to otherwise too many features.} For multi-classification and multi-label task we train separate one-vs-rest logistic regression models. We use l1-regularization with C=0.4. Note that one could probably increase logistic regression performance by tuning parameters and features more specifically to the investigated tasks. %

\textbf{Renyi Efficiency.} We record some obeservations: Using \fakesc{no} pre-tokenizer consistently got the highest Rényi Efficency and the \fakesc{ws} pre-tokenizer consistently got the lowest Rényi Efficency values, even though both were usually not among the best performing pre-tokenizers in the downstream tasks. 

\label{app:intrinsic_vocab-size} \textbf{Varying Vocabulary Size.} Corpus Token Count is sensitive to the vocabulary size of the tokenizer. A tokenizer with a vocabulary size of 128k will almost always have a lower Corpus Token Count than a tokenizer with the vocabulary size of 32k. Independent of how many tokens might be a better fit for a given corpus or task. Similarly, Rényi efficiency \cite{zouhar-etal-2023-tokenization} assumes the same vocabulary size to make the efficiency values comparable across tokenizers. Note that even when tokenizers have the same vocabulary size, the vocabulary coverage %
on the downstream corpus (i.e., the actual number of tokens that appear in the downstream corpus) might be smaller. For example, a tokenizer that was fitted on the Twitter corpus might include vocabulary that never appears in the original GLUE tasks. As a result Corpus Token Count and Rényi efficiency might be skewed for tokenizers that have a very low overlap in vocabulary with the downstream task corpus.

\clearpage
\newpage

\section{Modeling Results on Diverging Pre-training and Fitting Corpus} \label{app:webbook-pretraining}

\begin{figure}[t!]
    \centering
    \includegraphics[width=0.32\linewidth]{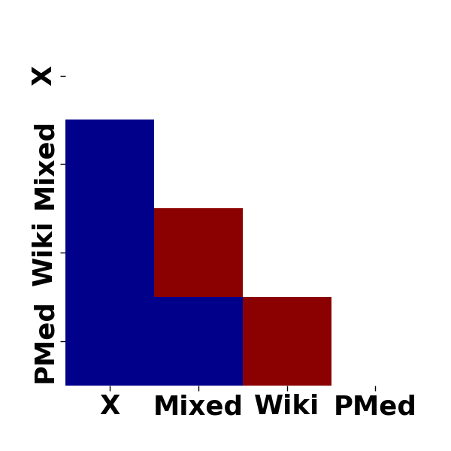}
    \includegraphics[width=0.32\linewidth]{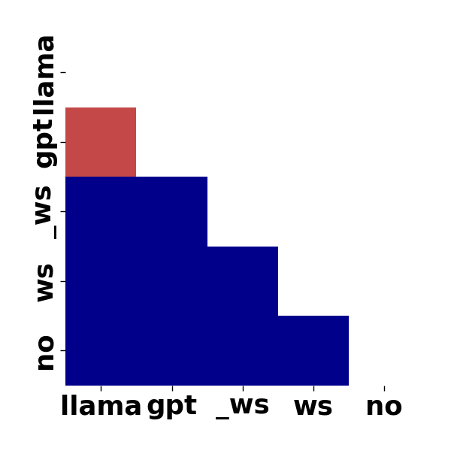}
    \includegraphics[width=0.32\linewidth]{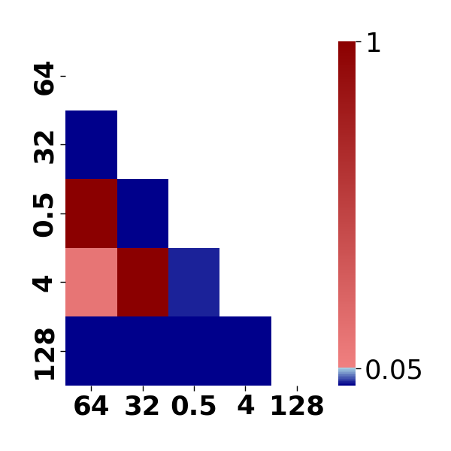}
    \includegraphics[width=0.32\linewidth]{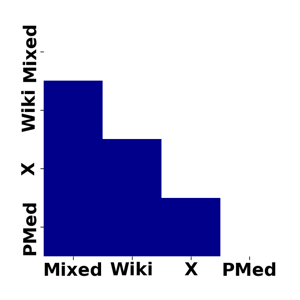}
    \includegraphics[width=0.32\linewidth]{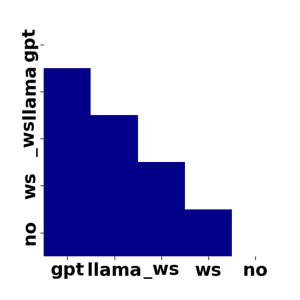}
    \includegraphics[width=0.32\linewidth]{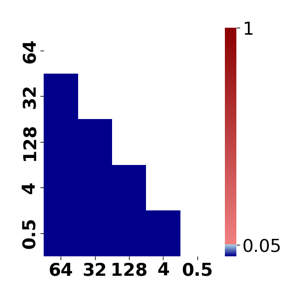}
    \caption{\textbf{Pairwise significance testing for tasks robust and sensitive to language variation with Mismatched Pre-training and Fitting Corpus.}
    We use the WebBook pre-training corpus and fit on the Miscellaneous, PubMed, Wikipedia and Twitter corpora. We use \cite{mcnemar1947note}'s test to test how different the BERT models trained with different tokenizers classify the tasks robust to language variation (first row) and tasks sensitive to language variation (second row). Tokenizers are sorted by mean performance. Blue colors show statistical significance, while red colors are above the 0.05 threshold.
    }
    \label{fig:webbook_sig}
\end{figure}

We experimented with diverging pre-training and fitting corpora. Specifically, for the pre-training corpus, we sampled 750 words from \textbf{WebBook} (\S\ref{app:modeling}). %
We expect WebBook corpus to show less spelling and syntactic variation than Miscellaneous used in the main paper. For the fitting corpus, we compared using PubMed, Wikipedia, Twitter and Miscellaneous. Further, we experimented with fitting on the Miscellaneous corpus but varying the pre-tokenizer and vocabulary size. %
See results in Table \ref{tab:webbook-GLUE-performance}, Table \ref{tab:webbook_Style-performance} and Figure \ref{fig:webbook_sig}. In contrast to the results in the main paper, the performances are not averaged over three seeds and should be considered preliminary.

For mismatched pre-training and fitting corpus, the vocabulary size needs to be higher for tasks requiring robustness to language variation. This is intuitive as more tokens during tokenizer fitting increase the likelihood of overlap with the pre-training corpus.

Further, the pre-tokenizers \fakesc{no}, \fakesc{ws} and \fakesc{\_ws} seem to have more trouble leveraging their tokens having been fitted on a corpus different from the pre-training corpus. This is intuitive as their tokens can be expected to be especially dependent on the fitting corpus. This is probably also the explanation for \fakesc{llama3} performing worse than \fakesc{gpt2} for tasks requiring sensitivity to language variation. \fakesc{gpt2} separates Unicode character categories the most and might thus have the largest overlap with the pre-training corpus. %

For mismatched pre-training and fitting corpus, the choice of fitting corpus seems more influential for tasks requiring robustness to language variation. Potentially the lack of overlap between the fitting and the pre-training corpus change what tokens can be leveraged effectively. Wikipedia performs surprisingly well for tasks that require sensitivity to language variation. Wikipedia is a corpus using a very standardized version of English. We theorize that this might help with recognizing deviations from that norm. %

\begin{table}[t!]
    \centering\small
    \begin{tabular}{ll | lll |l}
    \toprule
    & \textbf{Model} & \textbf{org} & \textbf{+typo} & \textbf{+dialect}  & \textbf{AVG}\\
    \midrule
     \multirow{3}{*}{\textbf{Corpus}} 
      & PMed 
        & 81.3 %
        & 69.4 %
        & 79.2 %
        & 76.6 \\ %
      & Wiki
        & {82.1} %
        & 68.9 %
        & 79.6 %
        & 76.8 %
        \\
      & Twitter 
        & \textbf{82.4} %
        & \textbf{70.4} %
        & \textbf{80.5} %
        & {\textbf{77.8}} %
        \\
      & Misc.
        & {81.6} %
        & 69.2 %
        & 79.9 %
        & 76.9 %
        \\ 
        \cmidrule{2-6}
    \multirow{4}{*}{\textbf{Pre-Tok}}
        & \fakesc{no} 
            & 72.3 %
            & 61.6 %
            & 70.9 %
            & 68.3 \\ %
        & \fakesc{ws} 
            & 79.5 %
            & 66.9 %
            & 78.1 %
            & 74.8 %
            \\ 
        & \fakesc{\_ws} 
            & 80.7 %
            & 68.6 %
            & 79.0 %
            & 76.1 %
            \\ 
        & \fakesc{llama3} 
            & \textbf{82.2} %
            & \textbf{69.4} %
            & {79.7} %
            & {\textbf{77.1}} %
            \\ 
        & \fakesc{gpt2} 
            & {81.6} %
            & 69.2 %
            & \textbf{79.9} %
            & {\textit{76.9}} %
            \\ 
        \cmidrule{2-6}
     \multirow{6}{*}{\textbf{Size}}
        & 500
            & 79.6 %
            & \textbf{72.8} %
            & 78.1 %
            & \textit{76.8} %
            \\ 
     & 4k 
            & 80.9 %
            & 70.8 %
            & 78.9 %
            & \textit{76.8} %
            \\ 
      & 32k & {81.6} %
            & 69.2 %
            & 79.9 %
            & 76.9 %
            \\ 
     & 64k 
            & \textbf{82.5} %
            & 69.3 %
            & \textbf{80.7 }%
            & {\textbf{77.5}} %
            \\ 
     & 128k 
            & {81.9}  %
            & 68.5 %
            & 79.9 %
            & 76.7 %
            \\ 
    \bottomrule
    \end{tabular}
    \caption{
        \textbf{Mismatched Pre-training and Fitting Corpus on Tasks Robust to Language Variation.} \label{tab:webbook-GLUE-performance}
        We use the WebBook pre-training corpus and fit on the Miscellaneous, PubMed, Wikipedia and Twitter corpora.
    }
\end{table}

\begin{table*}[t!]
\centering\small
    \begin{tabular}{ll|ccccc|c}
    \toprule
    & \textbf{Model} & \textbf{AV} (acc $\uparrow$) & \textbf{PAN} (acc $\uparrow$) & \textbf{CORE} (acc $\uparrow$) & \textbf{NUCLE} (F1 $\uparrow$) & \textbf{DIAL} (F1 $\uparrow$) & \textbf{Mean} \\ 
    \midrule
    \multirow{3}{*}{\textbf{Fitting Corpus}}  
        & PubMed 
            & 80.6 %
            & 65.5 %
            & 55.6 %
            & 23.8 %
            & 88.3 %
            & 62.8 %
            \\ 
        & Wikipedia 
            & 80.9 %
            & 67.2 %
            & {56.8} %
            & \textbf{24.8} %
            & {88.7} %
            & {63.7} %
            \\ 
        & Twitter 
            & 80.8 %
            & 64.7 %
            & {57.2} %
            & 23.5 %
            & \textbf{89.1} %
            & 63.0 %
            \\ 
        & Misc. 
            & \textbf{82.0} %
            & \textbf{68.3} %
            & \textbf{57.9} %
            & 24.1 %
            & {89.0} %
            & \textbf{64.3} %
            \\ 
    \cmidrule{2-8}
    \multirow{4}{*}{\textbf{Pre-Tokenizer}}  
        & \fakesc{no} 
            & 79.8 %
            & 52.4 %
            & 51.5 %
            & 16.9 %
            & 76.6 %
            & 55.4 %
            \\ 
        & \fakesc{ws} 
            & 74.9 %
            & 65.2 %
            & {55.7} %
            & 16.6 %
            & 85.3 %
            & 59.5 %
            \\ 
        & \fakesc{\_ws}
            & 80.4 %
            & 65.2 %
            & {56.8} %
            & 22.1 %
            & 87.9 %
            & 62.5 %
            \\ 
        & \fakesc{llama3}
            & 81.3 %
            & 66.9 %
            & {56.8} %
            & 23.4 %
            & \textbf{89.0} %
            & {63.5} %
            \\ 
        & \fakesc{gpt2}  
            & \textbf{82.0} %
            & \textbf{68.3} %
            & \textbf{57.9} %
            & \textbf{24.1} %
            & \textbf{89.0} %
            & {\textbf{64.3}} %
            \\ 
    \cmidrule{2-8}
    \multirow{6}{*}{\textbf{Vocabulary Size}}  
        & 500 
            & 77.6 %
            & 62.7 %
            & 53.2 %
            & 15.0 %
            & 86.6 %
            & 59.0 %
            \\ 
        & 4k  
            & 80.3 %
            & 65.1 %
            & 56.0 %
            & 21.3 %
            & 88.1 %
            & 62.2 %
            \\ 
        & 32k  
            & {82.0} %
            & \textbf{68.3} %
            & {57.9} %
            & 24.1 %
            & {89.0} %
            & \textbf{64.3} %
            \\ 
        & 64k 
            & 82.1 %
            & 67.2 %
            & \textbf{58.0} %
            & \textbf{25.4} %
            & 88.9 %
            & \textbf{64.3} %
            \\  
        & 128k 
            & \textbf{82.6} %
            & 66.6 %
            & 54.3 %
            & 24.1 %
            & \textbf{89.5} %
            & 63.4 %
            \\  
    \bottomrule
    \end{tabular}
\caption{
    \textbf{Mismatched Pre-training and Fitting Corpus on Tasks Sensitive to Language Variation.} 
    We use the WebBook pre-training corpus and fit on the Miscellaneous, PubMed, Wikipedia and Twitter corpora.
}
\label{tab:webbook_Style-performance}
\end{table*}

\clearpage
\newpage

\section{Intended Use and Licenses for used Datasets}

We discuss intended use and licenses for the datasets we re-used and created for this work.

\subsection{Datasets curated from different sources}

\textbf{The Pile.} We include several datasets extracted from \newcite{gao2020pile}'s The Pile. The Pile consists of newly collected datasets as well as datastes from other sources. There is no license information included the paper or original data release, but the Pile is described as ``open source language modelling data''\footnote{\url{https://pile.eleuther.ai/}} and, even though not explicitly stated, the intended use should be for open source language modelling and research. We use the Pile to access datasets from originally other sources: books from the 1919 Gutenberg project \cite{rae2019compressive}\footnote{Project Gutenberg consists mostly of public US ebooks, see \url{https://www.gutenberg.org/policy/permission.html}}, StackExchange\footnote{anonymized data shared with a cc-by-sa 4.0 license, see \url{https://archive.org/details/stackexchange}}, OpenSubtitles \cite{tiedemann-2016-finding}\footnote{No licensing or intended use information included. Originally extracted from \url{https://www.opensubtitles.org/}.}, Common Crawl\footnote{see Terms of Use: \url{https://commoncrawl.org/terms-of-use}} and DeepMind\footnote{No licensing information included in the paper.} \cite{saxton2019analysing}. We further use the following datasets collected by the authors of the Pile: GitHub,  OpenWebText2 and PubMed\footnote{See terms of the original dataset here: \url{https://ftp.ncbi.nlm.nih.gov/pubmed/baseline/README.txt}}.

\textbf{Reddit.} We use dataset originally downloaded from Pushshift \cite{baumgartner2020pushshift}. While the original release was public, and used in many research publications, Reddit updated their terms and the original Pushshift releases are not publicly accessible anymore. Reddit mentions at least partial support of academic research after agreeing to their terms of service.\footnote{\href{https://support.reddithelp.com/hc/en-us/articles/14945211791892-Developer-Platform-Accessing-Reddit-Data\#h_01H69EJ3EFY7G7HNV17ASH24KS}{Developer Platform \&{} Accessing Reddit Data}} 

\textbf{Ao3.} We downloaded a public release of Archive of Our Own from \url{https://archive.org/download/AO3_story_dump_continuing} in 2023. It did not include license nor intended use descriptions. The dataset was removed by now but should be re-creatable using tools like \url{https://github.com/nianeyna/ao3downloader}. Note that the license situation remains unclear. AO3 has a complex license where authors retain their rights and the website is granted a `a world-wide, royalty-free, nonexclusive license to make your Content available'. AO3 terms of service forbids use of fanfictions for commercial generative AI.\footnote{\url{https://archiveofourown.org/tos_faq}}

\textbf{Amazon Reviews.} The Amazon Reviews dataset \cite{hou2024bridging} was downloaded from \url{https://cseweb.ucsd.edu/~jmcauley/datasets/amazon_v2/}. While the dataset is publicly available, the license for the data remains with Amazon but the customers who wrote the reviews retain the copyright. There is no general site that provides guidance on the license and constraints for this data when used in the academic or research space. The guidelines for Amazon Services are noted here \url{https://www.amazon.com/gp/help/customer/display.html?nodeId=508088}  

\textbf{GoodReads.} GoodReads was publicly released with \newcite{wan2018item,wan-etal-2019-fine} for academic use. We downloaded it through \url{https://mengtingwan.github.io/data/goodreads}. The Goodreads license is available at \url{https://www.goodreads.com/about/terms}. The license includes the text `This license does not include any resale or commercial use of any part of the Service, or its contents; any collection and use of any book listings, descriptions, reviews or other material included in the Service; any derivative use of any part of the Service or its contents; any downloading, copying, or other use of account information for the benefit of any third party; or any use of data mining, robots, or similar data gathering and extraction tools.' 

\textbf{GMane.} Public mailing list emails collected from the gmane.io server, available at \url{https://webis.de/data/webis-gmane-19.html}. Released with \newcite{bevendorff-etal-2020-crawling}. Accessed through \url{https://zenodo.org/records/3766985} after submitting a request. No license stated but not publicly available without request. Terms of use are documented on the dataset website \url{https://zenodo.org/records/3766985}.

\textbf{Blogcorpus.} Released with \newcite{schler2006age}. The paper does not discuss license or intended use. Accessed through \url{https://www.kaggle.com/datasets/rtatman/blog-authorship-corpus}. \newcite{schler2006age} downloaded from \url{https://www.blogger.com/}.

\textbf{Bookcorpus3.} Originally released with \newcite{books1}. \newcite{bandy2021addressing} released a retrospective datasheet. \newcite{books1} did not discuss intended use or licenses. However, the license for the data can be expected to remain with the original book copyright holders, except in cases where the copyright has expired.

\textbf{NYTimes.} The dataset is publicly available at \url{https://www.kaggle.com/datasets/benjaminawd/new-york-times-articles-comments-2020} shared with a CC BY-NC-SA 4.0 license. However, in all likelihood the license still belongs to the NYTimes while the copyright remains with the commenter. Some details are availble here \href{https://help.nytimes.com/hc/en-us/articles/360039332111-The-New-York-Times-Content-Agreement}{https://help.nytimes.com/hc/en-us/articles/360039332111-The-New-York-Times-Content-Agreement}.

\textbf{Realnews.} Published with \newcite{zellers2019grover}. Downloaded from \url{https://github.com/rowanz/grover/tree/master/realnews}. License can be found at this \href{https://docs.google.com/forms/d/1LMAUeUtHNPXO9koyAIlDpvyKsLSYlrBj3rYhC30a7Ak/viewform?edit\_requested=true}{Google Docs Form}. It is intended only for research and education use and can not be distributed. 

\textbf{SFU-Socc.} Released with \newcite{kolhatkar2020sfu}. Downloaded from \url{https://github.com/sfu-discourse-lab/SOCC}. Shared with Creative Commons Attribution-NonCommercial-ShareAlike 4.0 International License.

\textbf{s2orc.} Released with \newcite{lo-etal-2020-s2orc} mentioning research and development as intended use. Downloaded through \url{https://github.com/allenai/s2orc/?tab=readme-ov-file}. License is given as ODC-By 1.0.

\textbf{YouTubeCommons.} Downloaded through \url{https://huggingface.co/datasets/PleIAs/YouTube-Commons}. Released with CC-By license. All transcripts are part of a video shared under a CC-By license.

\textbf{GLUE.} Released a collection of tasks with \newcite{wang-etal-2018-glue}. Downloaded from \url{https://huggingface.co/datasets/nyu-mll/glue}. GLUE is a common public dataset used to evaluate language models. We use MNLI \cite{williams-etal-2018-broad}, SST-2 \cite{socher-etal-2013-recursive}, QQP and QNLI.

\textbf{PAN.} This PAN 2024 dataset was extracted from Reddit \cite{ayele2024overview}. We downloaded the dataset from \url{https://zenodo.org/records/10677876}. Reddit's terms of use might apply.

\textbf{NUCLE.} We use the NUCLE 3.3 corpus 
\cite{dahlmeier-etal-2013-building}, downloaded from {\url{https://www.comp.nus.edu.sg/~nlp/corpora.html}} after submitting a request. It is available for for research purposes. License information can be found at \url{https://sterling8.d2.comp.nus.edu.sg/nucle_download/nucle.php} and does not allow for distribution of the corpus.

\textbf{CORE.}  Released with \cite{laippala2023register}, downloaded from \url{https://github.com/TurkuNLP/CORE-corpus}. It is released with a CC BY-SA 4.0 license.

\subsection{Datasets collected by us}

\textbf{Twitter.} Sampled in 2023 with Twitter research API access using the Decahose sampling stream. Distribution of tweet texts was not granted under research access. Intended use was academic research.

\section{Personally Identifying Info Or Offensive Content in Datasets}

Some of the used datasets can be expected to include personally identifying information or offensive content. We did not take steps to remove identifiable cues or offensive content. This was out of scope for the extensive amount of datasets used. We hope that the effect is negligible as for all datasets, except for Twitter, datasets were already publicly accessible. We acknowledge that re-distributing it might, however, make it more widely accessible. We do not release the Twitter dataset publicly.

\section{Model Size and Budget}

We used single A100s to run modeling. We pre-trained 24 distinct BERT models for our main experiments (taking less than 360 GPU hours), and fine-tuned each model for all evaluation tasks ($\approx24 * (6h*3$ [GLUE tasks] + $3h$ [tasks requiring sensitivity language variation]$)=24*21h=504h$). 

\section{Use of AI Assistants}

We used ChatGPT and GitHub Copilot for coding, to look up commands and sporadically to generate individual functions. %
Generated functions were tested w.r.t. expected behavior. We used AI assistants for rephrasing and grammatical error correction in writing.

\end{document}